%% file: main.tex
\documentclass{article}

\usepackage[preprint]{neurips_2026}

\usepackage[utf8]{inputenc} % allow utf-8 input
\usepackage[T1]{fontenc}    % use 8-bit T1 fonts
\usepackage{hyperref}       % hyperlinks
\usepackage{url}            % simple URL typesetting
\usepackage{booktabs}       % professional-quality tables
\usepackage{amsfonts}       % blackboard math symbols
\usepackage{nicefrac}       % compact symbols for 1/2, etc.
\usepackage{microtype}      % microtypography
\usepackage{xcolor}         % colors

\usepackage{amsmath}
\usepackage{graphicx}
\usepackage{array}
\usepackage{multirow}
\usepackage{tikz}
\usepackage{pifont}
\usepackage[table,dvipsnames]{xcolor}
\usepackage{tabularx, colortbl, multirow}
\usepackage[most]{tcolorbox}
\usepackage{algorithm}
\usepackage{algorithmic}
\usepackage{fvextra}
\usepackage{marvosym}

\hypersetup{
colorlinks=true,
linkcolor=red,
}

\usetikzlibrary{arrows.meta, positioning, fit, calc, shapes.geometric, decorations.pathreplacing}

\definecolor{stainred}{RGB}{205, 54, 65}
\definecolor{stainblue}{RGB}{36, 114, 178}
\definecolor{staingreen}{RGB}{40, 145, 92}
\definecolor{stainorange}{RGB}{224, 135, 37}
\definecolor{softgray}{RGB}{246, 247, 249}
\definecolor{linegray}{RGB}{118, 125, 135}
\newcommand{\ind}{\mathbf{1}}

\title{StainFlow: Entity-Stain Tracking and Evidence Linking for Process Rewards in GUI Agents}

\author{
  Haojie Hao$^{1}$, Longkun Hao$^{1}$, Yihang Lou$^{2,\text{\Letter}}$, Yan Bai$^{2}$, Zhenyang Li$^{3}$, \textbf{Zhichao Yang}$^{1}$,\\
  \textbf{Dongshuo Huang}$^{4}$, \textbf{Hongyu Lin}$^{5}$, \textbf{Lanqing Hong}$^{6}$, \textbf{Jiakai Wang}$^{7}$, \textbf{Xianglong Liu}$^{7}$ \\
  $^{1}$Beihang University\quad
  $^{2}$Peking University\quad
  $^{3}$Renmin University of China\\
  $^{4}$Northwestern Polytechnical University\quad
  $^{5}$Institute of Software, Chinese Academy of Sciences\\
  $^{6}$National University of Singapore\quad
  $^{7}$Zhongguancun Laboratory\\
  \texttt{\small haojiehao@buaa.edu.cn} \quad \texttt{\small yihanglou@pku.edu.cn}
}

\begin{document}

\maketitle

\begin{abstract}

Reinforcement Learning (RL) has become a promising approach for improving GUI Agents in long-horizon, stochastic digital environments, but trajectory-level success feedback is too sparse to provide reliable credit assignment for intermediate exploration steps.
To mitigate this issue, recent studies introduce Process Reward Models (PRMs), which provide finer-grained training feedback through global milestone verification or local step-level evaluation.
However, these methods still suffer from two level-specific limitations: global milestone decomposition is subjective and singular, making it difficult to accommodate the multiple valid execution paths in real GUI tasks, while fixed local judging windows may miss long-range key evidence or dilute the decision signal with irrelevant frames.
Inspired by stain-tracing mechanisms in network flow analysis, we propose StainFlow, an entity-stain-flow process reward model for GUI Agents.
To reduce the subjectivity of global partitioning, we introduce the Global Entity Stain Tracking module, which extracts visually verifiable task entities and tracks how their stain concentrations and states evolve along the trajectory, allowing task phases to be objectively separated by changes in the entity evidence flow.
To improve the accuracy of local verification, we introduce the Local Stain Evidence Linking module. Centered on the triggering entities of each candidate key node, it retrieves relevant steps based on their stain concentrations and state changes, and dynamically constructs high-density evidence windows for verifying true key nodes.
Extensive experiments on AndroidWorld and OGRBench show that StainFlow relatively improves online RL success by 3.2\% and trajectory completion judgment accuracy by 1.8\%.

\end{abstract}

\section{Introduction}

Recent advances in multimodal large language models (MLLMs)~\cite{bai2025qwen3vl,qwen3.5,singh2025openai,googledeepmind2026gemini31pro} have made graphical user interface (GUI) agents~\cite{wang2025uitars2,xu2026mobileagentv35,gu2025uivenus,zhou2025maiui,xue2026evocua,wang2025opencua} a promising paradigm for automating complex digital tasks.
With native visual reasoning, GUI Agents can interpret screenshots, locate interface elements, and execute operations.
Building on this foundation, online Reinforcement Learning (RL)~\cite{bai2024digirl,wang2024distrl,lu2025arpo,shi2025mobilegui,luo2025gui} further improves GUI Agents through interactive feedback, strengthening their navigation and recovery abilities in stochastic digital environments.
However, RL depends on reward quality.
Trajectory-level success signals in GUI tasks are sparse and delayed~\cite{feng2025group,kuang2025tim,xi2026agentprm}, making it difficult to reward progress in failed trajectories or penalize redundant detours.

\begin{figure}[t]
    \centering
    \includegraphics[width=\linewidth]{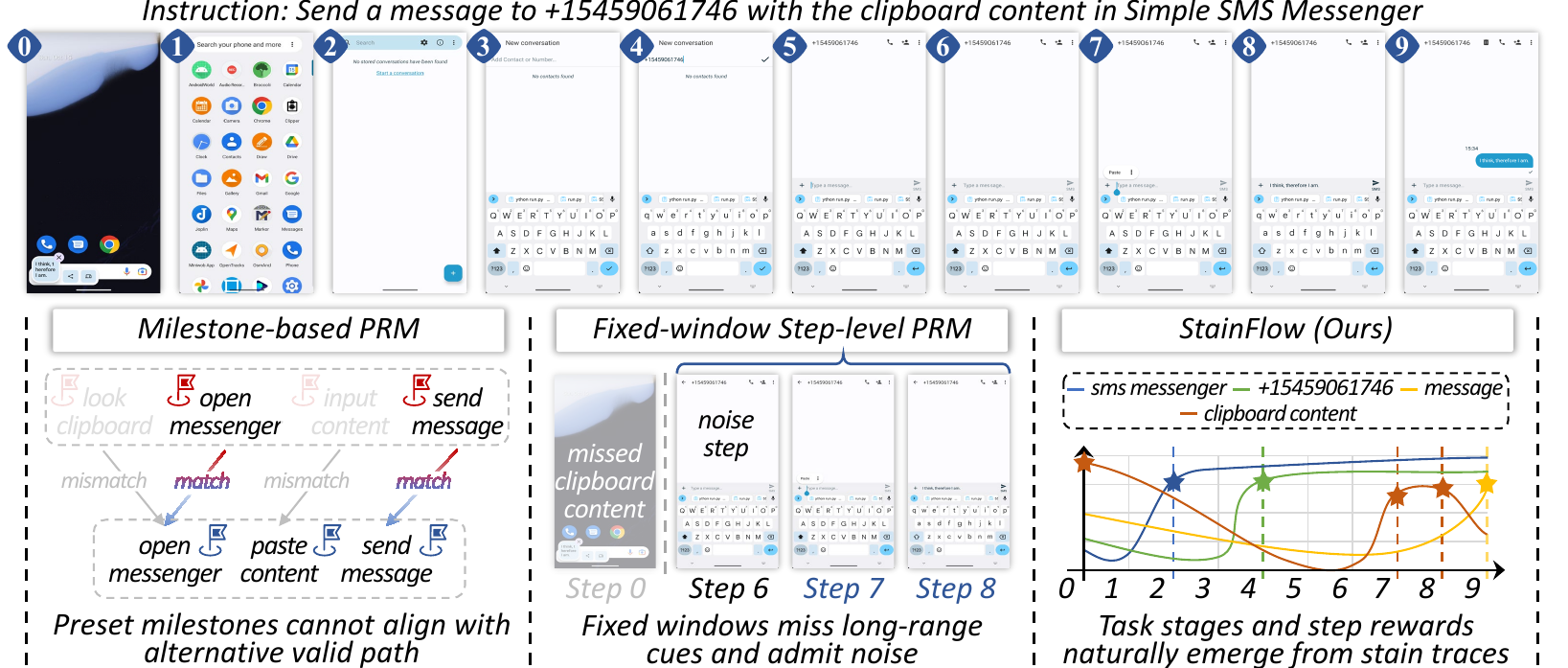}
    \caption{Our proposed StainFlow dynamically and objectively partitions task stages and introduces flexible evidence windows for key-node verification, thereby providing more accurate step rewards.} 
    \label{fig:intro}
\end{figure}

To provide denser supervision, recent studies introduce Process Reward Models (PRMs) for intermediate steps~\cite{choudhury2025process}.
One line performs global milestone verification by decomposing tasks into stage-level subgoals and checking trajectory segments, as in ADMIRE~\cite{zheng2026adaptive} and OS-Themis~\cite{li2026themis}.
Another line performs local step-level evaluation, where GUI-Critic-R1~\cite{wanyan2025guicritic} and GUI-PRA~\cite{xiong2025gui} judge action value from the current screenshot, action, and short context.
The former offers interpretable stage rewards, while the latter gives immediate action feedback, forming the main design space for GUI PRMs.

Despite their progress, existing PRMs still suffer from two hierarchical limitations in global task partitioning and local evidence selection.
\ding{182} At the global level, milestone-based methods usually rely on a predefined or generated subtask sequence, compressing multi-solution GUI tasks into a subjective and singular execution path. When a GUI Agent reaches the same goal through alternative routes such as search, share sheets, or histories, the predefined milestones may under-credit real progress or reward planned steps that are irrelevant to the executed trajectory.
\ding{183} At the local level, step-level evaluation methods usually rely on fixed-length context windows, constraining key evidence to a uniform and rigid temporal range. When reliable judgment depends on long-range cues such as first target discovery, state updates, or delayed confirmation, short windows may miss critical evidence, while long windows introduce irrelevant frames and dilute the decision signal.

To address these limitations, inspired by stain-tracing mechanisms in network flow analysis~\cite{savage2001network,muthuprasanna2006coloring}, we model GUI trajectories and key entities as network paths and flow information, proposing StainFlow, an entity-stain-flow PRM that converts visually verifiable task entities into dynamic stain signals along the trajectory.
These signals let global task stages emerge from entity evidence flow and help local key steps aggregate evidence around relevant entities.
To reduce the subjectivity of global partitioning, we introduce \textbf{Global Entity Stain Tracking}. This module extracts key entities that can be observed and verified from the interface, and dynamically adjusts their stain concentrations based on visibility and state changes, which causes the concentrations to rise, remain, or decay. It then recalls candidate key nodes from the high concentration, persistence, and abrupt changes of entity stains, thereby partitioning task stages in a natural and evidence-driven manner.
To improve the accuracy of local verification, we propose \textbf{Local Stain Evidence Linking}. Centered on the triggering entities of each candidate node, this module traces their full-trajectory stain histories to retrieve steps with high concentration, sharp changes, or temporal proximity. It therefore constructs adaptive, evidence-dense local windows for different candidates as compact evidence contexts that concentrate the key information of triggering entities, supporting accurate key-node verification.
Finally, the step reward combines continuous entity-stain concentrations with discrete key-node rewards. This credits genuine intermediate progress without relying on subjective milestones or fixed windows, minimizing the interference of irrelevant steps on policy updates. Our contributions are summarized as follows:

\begin{itemize}
  \item We propose StainFlow, an entity-stain-flow PRM inspired by stain-tracing mechanisms in network flow analysis. By converting task-entity visibility and state changes into evolving stain signals, it provides objective, fine-grained rewards for long-horizon GUI RL.
  \item We introduce Global Entity Stain Tracking and Local Stain Evidence Linking. The former uses entity stain dynamics to discover task stages from the executed trajectory, reducing the influence of subjective milestone planning; the latter retrieves dynamic evidence windows around triggering entities, mitigating evidence loss or dilution caused by fixed local contexts.
  \item Extensive experiments on AndroidWorld and OGRBench show that StainFlow relatively improves online RL success by 3.2\% and trajectory completion judgment accuracy by 1.8\%, effectively enhancing reward-guided policy learning and trajectory assessment.
\end{itemize}

\section{Related Work}

\paragraph{GUI Agents and Reinforcement Learning.}
Recent MLLMs~\cite{bai2025qwen3vl,qwen3.5,singh2025openai,googledeepmind2026gemini31pro} have moved GUI Agents from offline visual prediction to real interface interaction, with SeeClick~\cite{cheng2024seeclick}, CogAgent~\cite{hong2024cogagent}, and AssistGUI~\cite{gao2024assistgui} improving element grounding, GUI understanding, and action generation.
As tasks become longer and more stochastic, DigiRL~\cite{bai2024digirl}, ZeroGUI~\cite{yang2025zerogui}, and ARPO~\cite{lu2025arpo} use online Reinforcement Learning to optimize device-control policies from interaction data.
Yet these pipelines still rely heavily on trajectory-level success signals from rule-based verification, which are sparse, delayed, and difficult to scale in long-horizon GUI tasks, motivating finer-grained reward modeling.

\paragraph{Process Rewards for GUI Agents.}
PRMs mitigate credit assignment by supervising intermediate steps in long-horizon tasks.
Existing GUI PRMs can be grouped into global and local methods.
Global methods estimate progress from trajectories or task stages, including GUI-PRA~\cite{xiong2025gui}, ProgRM~\cite{zhang2025progrm}, and ADMIRE~\cite{zheng2026adaptive}.
They improve feedback density and interpretability, but may bind multi-solution tasks to a particular progress definition or subtask chain.
OS-Themis~\cite{li2026themis} mitigates reliance on a single milestone chain by iterating milestones through an agent pipeline, but introduces substantial inference cost.
Local methods such as GUI-Shepherd~\cite{chen2025guishepherd}, OS-Oracle~\cite{wu2025osoracle}, and GUI-Critic-R1~\cite{wanyan2025guicritic} judge actions or short contexts more directly, but often miss long-range visual evidence for executed steps.
Unlike milestone-based or step-level PRMs, StainFlow tracks entity stains along trajectories and improves global partitioning and local evidence organization through objective entity-stain flows.

\paragraph{Reward Benchmarks and Evaluation.}
GUI Agent capability benchmarks such as AndroidWorld~\cite{rawles2025androidworld}, OSWorld~\cite{xie2024osworld}, and WebArena~\cite{zhou2024webarena} measure post-execution success across mobile, desktop, and web tasks.
Reward-accuracy benchmarks mostly focus on trajectory-level judgment, as in OGRBench~\cite{li2026themis} and AgentRewardBench~\cite{lu2025agentrewardbench}.
Process-reward benchmarks such as OS-Oracle~\cite{wu2025osoracle} and GUI-Critic-Test~\cite{wanyan2025guicritic} usually reduce evaluation to step correctness, candidate preference, or local critic decisions, leaving whole-trajectory step value under-specified.
Since step value depends on later states, alternative paths, and task semantics, GUI PRMs should be evaluated by whether rewards improve downstream RL training, trajectory filtering, and trained-agent capability.

\section{Approach}

\subsection{Problem Definition}

Given instruction $g$, we describe GUI task as a goal-conditioned partially observable process~\cite{wang2024gui}. The interface evolves after actions, while the agent observes only screenshots and past actions.
Let $o_t$ be the screenshot at step $t$, $a_t$ the following action, and $h_t$ the visible history before action selection:
\begin{equation}
  h_t=(o_0,a_0,\ldots,o_t),\quad
  a_t\sim\pi_\theta(\cdot\mid h_t,g),\quad
  \tau=(o_0,a_0,o_1,a_1,\ldots,o_{T-1},a_{T-1},o_T).
  \label{eq:pomdp}
\end{equation}
In Eq.~\ref{eq:pomdp}, $\pi_\theta$ is the GUI policy model parameterized by $\theta$, $\tau$ is the executed trajectory, and $T$ is the terminal observation index.
The rewards for GUI tasks can be divided into trajectory-level and step-level rewards.
Trajectory-level rewards are usually available after execution from environment or task verifiers that check final task completion~\cite{cui2026agentic,dai2025prore}.
Step-level process value has no such direct feedback: useful actions may be verified only later, while redundant actions can still appear in successful trajectories.
Thus, the GUI process reward model relies on the full interaction trajectory to retrospectively evaluate the value of each step and produce dense, continuous per-step rewards:
\begin{equation}
  R(\tau)\in\{0,1\},\quad
  (r_0,\ldots,r_{T-1},\hat{y})=\phi(g,\tau),
  \quad r_t\in[0,1].
  \label{eq:prm_definition}
\end{equation}
Here $R(\tau)$ is the trajectory outcome reward, $\phi$ denotes the PRM, $\hat{y}$ is an optional completion prediction, and $r_t$ measures the value of the executed step at time $t$.
The challenge is assigning credit from trajectory context without direct step-level environment labels.

\begin{figure}[t]
    \centering
    \includegraphics[width=\linewidth]{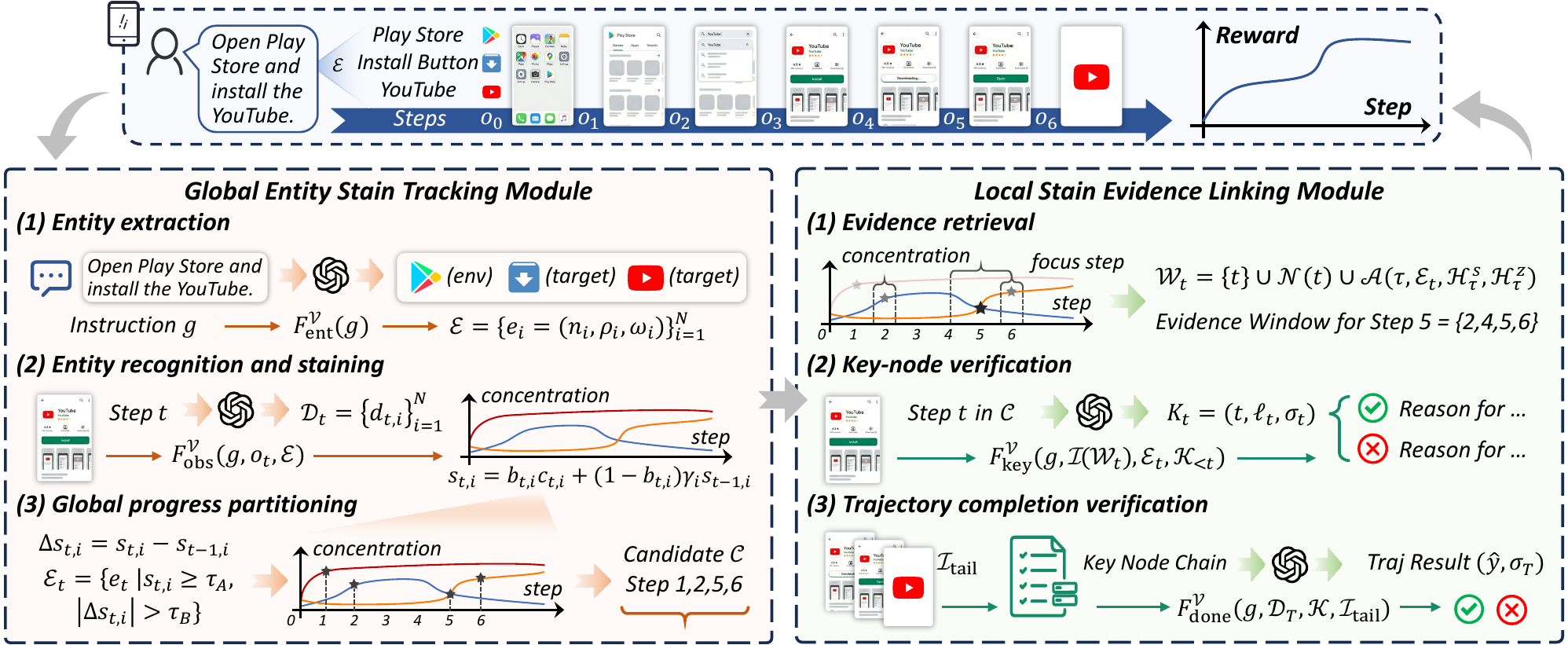}
    \caption{Overall workflow of StainFlow, where Global Entity Stain Tracking and Local Stain Evidence Linking enable more objective and fine-grained step reward assignment.} 
    \label{fig:framework}
\end{figure}

\subsection{Framework Overview}

To address subjective global milestone decomposition and rigid local windows, we formulate GUI PRMs as organizing the flow, recording, and retrieval of trajectory evidence.
Inspired by traffic coloring, recording, and traceback mechanisms in network security~\cite{savage2001network,muthuprasanna2006coloring}, StainFlow treats visually verifiable task entities as stain carriers and updates their visibility, states, and stain concentrations along the executed trajectory; these signals rise, persist, shift, or decay as entity evidence appears, persists, transfers, or disappears.
At the global level, valid GUI paths may vary, but progress usually manifests as state changes of task-relevant entities.
\textbf{Global Entity Stain Tracking} extracts visually verifiable entities, grounds their states, updates stain concentrations, and recalls candidate key steps from state transitions and stain dynamics, allowing task stages to emerge from the executed trajectory.
At the local level, evidence for judging a step may be scattered across distant steps.
\textbf{Local Stain Evidence Linking} retrieves support steps around the candidate's triggering entities, including high-stain, abrupt-change, and state-transition steps, and verifies true key nodes with an adaptive evidence window.
The resulting entity stains and verified key nodes are converted into step rewards for RL training. The overall framework is illustrated in Figure~\ref{fig:framework}.

\subsection{Global Entity Stain Tracking}

The goal of Global Entity Stain Tracking is to replace subjective milestone planning with evidence-driven progress discovery.
The module first identifies visually verifiable task entities, then lets the trajectory itself reveal where progress occurs through entity-state and stain dynamics.

\paragraph{Entity extraction.}

Given instruction $g$, the entity extraction function $F_{\mathrm{ent}}^{\mathcal{V}}$ outputs a set of screenshot-checkable task entities:
\begin{equation}
  \mathcal{E}=\{e_i=(n_i,\rho_i,\omega_i)\}_{i=1}^{N}=F_{\mathrm{ent}}^{\mathcal{V}}(g).
  \label{eq:entity_grounding}
\end{equation}
Here we use an auxiliary VLM $\mathcal{V}$ to implement $F_{\mathrm{ent}}^{\mathcal{V}}$, the input is the instruction $g$, and the output is the entity set $\mathcal{E}$.
The scalar $N$ is the number of entities, $n_i$ is an entity name, $\rho_i$ is a checkable rule for deciding whether the entity appears in a step screenshot, and $\omega_i$ stores attributes such as importance and persistence.
These entities establish the objective visual cues to be followed along the trajectory.

\paragraph{Entity recognition and staining.}

At each step $t$, the entity observation function $F_{\mathrm{obs}}^{\mathcal{V}}$ parses entity visual states from the instruction $g$, screenshot $o_t$, and entity set $\mathcal{E}$:
\begin{equation}
  \mathcal{D}_t=\{d_{t,i}\}_{i=1}^{N}=F_{\mathrm{obs}}^{\mathcal{V}}(g,o_t,\mathcal{E}),\quad
  d_{t,i}=(b_{t,i},c_{t,i},z_{t,i},v_{t,i}).
  \label{eq:entity_observation}
\end{equation}
The output $\mathcal{D}_t$ is the entity observation list at step $t$; each $d_{t,i}$ corresponds to entity $e_i$ and contains the visual-evidence flag $b_{t,i}$, confidence $c_{t,i}$, state-change description $z_{t,i}$, and current entity state $v_{t,i}$. If a new task-relevant entity is observed, it is added to $\mathcal{E}$ for later tracking.
Based on $d_{t,i}$, entity $i$ receives a stain concentration $s_{t,i}\in[0,1]$ after step $t$: if it appears in the current screenshot, the stain is set to the recognition confidence; otherwise, it decays by an entity-specific factor $\gamma_i$:
\begin{equation}
  s_{t,i}=b_{t,i}c_{t,i}+(1-b_{t,i})\gamma_i s_{t-1,i},\quad
  \Delta s_{t,i}=s_{t,i}-s_{t-1,i}.
  \label{eq:stain_update}
\end{equation}
Here $\gamma_i\in[0,1]$ is the discount factor specified by entity attributes $\omega_i$, and $\Delta s_{t,i}$ is the stain change at step $t$.
This update keeps recently observed entities at high concentration while gradually fading evidence for entities that have disappeared from the trajectory.

\paragraph{Global progress partitioning.}

Global task stages are partitioned by the joint evolution of entity stains and states, rather than fixed milestones or local judgments.
If any entity $e_i$ at step $t$ satisfies $s_{t,i}\ge\tau_A$ or $|\Delta s_{t,i}|>\tau_B$, step $t$ is recalled as a candidate key node, and all candidates form $\mathcal{C}$.
Here $\tau_A$ and $\tau_B$ are the stain and change thresholds.
For each $t\in\mathcal{C}$, the responsible entities are recorded as $\mathcal{E}_t\subseteq\mathcal{E}$ and passed to Local Stain Evidence Linking for stage-progress verification.
Thus, global stages arise from executed entity-evidence dynamics instead of a pre-specified milestone chain.

\subsection{Local Stain Evidence Linking}

Local Stain Evidence Linking verifies high-recall candidates with entity-centered evidence.
Instead of fixed windows, it retrieves evidence from the stain histories of triggering entities across the full trajectory, then uses the verified key nodes to judge final trajectory completion.

\paragraph{Evidence retrieval.}

For a candidate node $t$, the evidence window contains the candidate itself, a small local neighborhood, and support steps linked to its triggering entities $\mathcal{E}_t$ across the trajectory:
\begin{equation}
  \begin{aligned}
    \mathcal{W}_t = \{t\}\cup\mathcal{N}(t)\cup\mathcal{A}(\tau,\mathcal{E}_t,\mathcal{H}^{s}_\tau,\mathcal{H}^{z}_\tau)
  \end{aligned}
  \label{eq:flex_window}
\end{equation}
\begin{equation}
  \begin{aligned}
    \mathcal{A}(\tau,\mathcal{E}_t,\mathcal{H}^{s}_\tau,\mathcal{H}^{z}_\tau) = \{\,j \mid \exists e_i\in\mathcal{E}_t, s_{j,i}\ge\tau_A \vee |\Delta s_{j,i}|>\tau_B\,\}.
  \end{aligned}
  \label{eq:operator_A}
\end{equation}
Here $\mathcal{W}_t$ is the evidence window for candidate $t$, $\mathcal{N}(t)$ is a small local neighborhood, and $\mathcal{A}$ is the stain-association operator that recalls support steps related to triggering entities across the trajectory.
The histories $\mathcal{H}^{s}_\tau=\{s_{j,i}\}$ and $\mathcal{H}^{z}_\tau=\{z_{j,i}\}$ collect trajectory-wide entity stains and state changes.
Unlike fixed local context, $\mathcal{W}_t$ can include distant steps with concentrated triggering-entity evidence.

\paragraph{Key-node verification.}

A step is verified as a key node if it introduces a new task-relevant state change. Key-node verification produces a record from instruction $g$, composite evidence from $\mathcal{W}_t$, triggering entities $\mathcal{E}_t$, and the accepted key-node history $\mathcal{K}_{<t}$:
\begin{equation}
  K_t
  =
  (t,\ell_t,\sigma_t)
  =
  F_{\mathrm{key}}^{\mathcal{V}}\!\left(
    g,\mathcal{I}(\mathcal{W}_t),\mathcal{E}_t,\mathcal{K}_{<t}
  \right).
  \label{eq:key_verification}
\end{equation}
Here $K_t$ is the verification record for candidate $t$, containing the step index $t$, a binary key-node decision $\ell_t$, and a progress summary $\sigma_t$.
$\mathcal{I}(\mathcal{W}_t)$ denotes the composite visual evidence built from the selected screenshots.
A candidate is accepted only when the evidence establishes a new task-relevant fact that was not already covered by earlier key nodes.
The accepted nodes form $\mathcal{K}=\{K_t:t\in\mathcal{C},\ell_t=1\}$, an interpretable skeleton that filters redundant, failed, or unsupported candidates.

\paragraph{Trajectory completion verification.}

After all candidate nodes are verified and the key-node chain is built, the completion verification function $F_{\mathrm{done}}^{\mathcal{V}}$ takes the instruction $g$, final entity snapshot $\mathcal{D}_T$, key-node chain $\mathcal{K}$, and tail screenshot evidence $\mathcal{I}_{\mathrm{tail}}$ as inputs:
\begin{equation}
  (\hat{y},\sigma_T)
  =
  F_{\mathrm{done}}^{\mathcal{V}}\!\left(
    g,\mathcal{D}_T,\mathcal{K},\mathcal{I}_{\mathrm{tail}}
  \right).
  \label{eq:completion_verification}
\end{equation}
The outputs are the trajectory completion prediction $\hat{y}$ and final summary $\sigma_T$, which connect global entity states with the local key-node chain for inference-time trajectory completion judgment.

\subsection{Reward Assignment and Policy Update}

Effective credit assignment unifies continuous progress signals for smooth learning with discrete key events for precise allocation. StainFlow achieves this by coupling a stain component for dense, adaptive feedback with key-node verification to reward critical progress~\cite{setlur2024rewarding,wang2024math,peng2025agentic}.
The continuous term $p_t$ aggregates entity stain concentrations at step $t$, where $w_i$ is the importance weight associated with $\omega_i$.
The discrete term $k_t$ equals $1$ when a verified key node in $\mathcal{K}$ occurs at step $t$.
The step reward $\hat{r}_t$ combines the continuous stain term and the discrete key-node term:
\begin{equation}
  p_t=\frac{\sum_{i=1}^{N} w_i s_{t,i}}{\sum_{i=1}^{N} w_i},\quad
  \hat{r}_t=\lambda_r p_t+(1-\lambda_r)k_t.
  \label{eq:step_reward}
\end{equation}
Here $\lambda_r$ controls the mixture between stain evidence and key-node evidence.
The stain term rewards useful intermediate evidence, while the key-node term concentrates credit on verified progress.

For online RL, we follow the critic-free group optimization style of GRPO~\cite{shao2024deepseekmath} and the process-outcome alignment of PRPO~\cite{ding2026prpo}.
For each instruction $g$, the old policy samples a trajectory group $\mathcal{B}_g=\{\tau^{(m)}\}_{m=1}^{M}$.
Rather than assuming a cross-task homogeneous process-score distribution, we estimate $\mu_P$ and $\sigma_P$ from all step rewards in this instruction group.
The advantage is defined as:
\begin{equation}
  A_t^{(m)}
  =\frac{R^{(m)}-\mu_R}{\sigma_R+\epsilon}
  +\eta\frac{\hat{r}^{(m)}_t-\mu_P}{\sigma_P+\epsilon}.
  \label{eq:final_advantage}
\end{equation}
Here $A_t^{(m)}$ is the final advantage for step $t$ in trajectory $m$, $R^{(m)}=R(\tau^{(m)})$ is the trajectory outcome reward, $T_m$ is the length of that trajectory, $\eta$ is the advantage weight, and $\epsilon$ is a numerical stabilizer.
The statistics $\mu_R,\sigma_R$ are computed from group outcome rewards, while $\mu_P,\sigma_P$ are computed from $\{\hat{r}^{(m)}_t:1\leq m\leq M,0\leq t < T_m\}$, so that step-score normalization adapts to the candidate trajectories of the same instruction. The final training loss is defined as:
\begin{equation}
  \mathcal{L}(\theta)
  =
  \frac{1}{\sum_{m=1}^{M} T_m}
  \sum_{m=1}^{M}\sum_{t=0}^{T_m-1}
  \min\!\left(
    \rho_t^{(m)}A_t^{(m)},
    \operatorname{clip}\!\left(\rho_t^{(m)},1-\epsilon_c,1+\epsilon_c\right)A_t^{(m)}
  \right)
  +\beta\,\mathrm{KL}(\theta).
  \label{eq:rl_loss}
\end{equation}
\begin{equation}
  \begin{aligned}
  \rho_t^{(m)}
  &=
  \frac{\pi_\theta(a_t^{(m)}\mid h_t^{(m)},g)}
       {\pi_{\mathrm{old}}(a_t^{(m)}\mid h_t^{(m)},g)},
  \;\;
  \mathrm{KL}(\theta)
  =
  \mathrm{KL}\!\left(\pi_\theta\,\|\,\pi_{\mathrm{ref}}\right).
  \end{aligned}
  \label{eq:rl_terms}
\end{equation}
Here $\rho_t^{(m)}$ is the probability ratio between the new and old policies, $\mathcal{D}_{\mathrm{KL}}(\theta)$ is the reference-policy KL term, $\sum_{m=1}^{M}T_m$ is the number of optimized steps in the instruction group, $\epsilon_c$ is the clipping radius, $\beta$ is the KL penalty weight, and $\pi_{\mathrm{ref}}$ is the reference policy.
This loss assigns each executed GUI step its own advantage while preserving trajectory-level reward stability.

\section{Experiments}
\label{sec:experiment}

We conduct online RL experiments to verify whether StainFlow's step rewards improve GUI Agent training, and conduct inference-time trajectory completion judgment to examine whether entity-stain evidence and key-node chains improve full-trajectory assessment.
Finally, we perform ablation and analysis experiments to further analyze the roles of each module and key design choice.

\subsection{Experimental Settings}
\label{sec:Experiment_Settings}

\paragraph{Datasets.}
We evaluate on AndroidWorld~\cite{rawles2025androidworld} and OGRBench~\cite{li2026themis}.
AndroidWorld is a GUI Agent benchmark over real Android applications, and we use it to evaluate online RL training.
The training set is generated from AndroidWorld task templates with random seeds different from the validation set, yielding 928 trajectories.
OGRBench is a cross-platform GUI trajectory-level reward benchmark, and we use it for inference-time trajectory completion judgment.

\paragraph{Models.}
For online RL, the trained GUI policy model is Qwen3-VL-8B~\cite{bai2025qwen3vl}, and we use Qwen3.5-VL-9B~\cite{qwen3.5} and Qwen3.5-VL-27B~\cite{qwen3.5} as auxiliary verifier models.
For inference-time trajectory completion judgment, we evaluate a broader set of multimodal models, including Qwen3-VL-8B~\cite{bai2025qwen3vl}, Qwen3.5-VL-9B~\cite{qwen3.5}, Qwen3.5-VL-27B~\cite{qwen3.5}, GPT-5~\cite{singh2025openai}, and Gemini-3-Flash~\cite{googledeepmind2026gemini31pro}.

\paragraph{Baselines.}
For online RL, we compare with GUI-Critic-R1~\cite{wanyan2025guicritic}, ADMIRE~\cite{zheng2026adaptive}, and OS-Themis~\cite{li2026themis}, covering stepwise and milestone-style PRMs.
For GUI-Critic-R1, steps judged as correct and incorrect receive rewards of 1 and 0.
For ADMIRE and OS-Themis, milestone and non-milestone steps receive rewards of 1 and 0.
For inference-time trajectory completion judgment, we compare with DigiRL~\cite{bai2024digirl}, ZeroGUI~\cite{yang2025zerogui}, and OS-Themis.

\paragraph{Metrics.}
For online RL, we report AndroidWorld success rate, average step reward on successful trajectories, average step reward on failed trajectories, their reward gap, and average execution steps.
For inference-time trajectory completion judgment, following OS-Themis, we report Acc. and F1 on the five OGRBench subsets, together with overall Acc., Prec., Recall, and F1.

\paragraph{Implementation Details.}
For online RL, each method is trained for 5 epochs with batch size 64, GRPO group size 8, learning rate $1\times10^{-5}$, and $\eta=0.5$.
For StainFlow, entities whose attribute $\omega_i$ indicates persistence, such as app-level execution environments, use stain decay $\gamma_i=0.8$, and other entities use $\gamma_i=0.5$.
The high-stain threshold is $\tau_A=0.9$, the stain-change threshold is $\tau_B=0$, the neighborhood $\mathcal{N}(t)$ includes one step before and after the candidate, the tail evidence $\mathcal{I}_{\mathrm{tail}}$ uses the last three steps, and $\lambda_r=0.4$.
Implementation details for other methods are deferred to the appendix.

\subsection{Online RL Training Results}

We conduct online RL on AndroidWorld, with results shown in Table~\ref{tab:androidworld}.
Overall, StainFlow achieves the highest success rate and a more reasonable step-reward distribution: successful trajectories receive high but non-saturated rewards, failed trajectories retain credit for partial progress, and the two groups keep a clear gap.
This shows that Global Entity Stain Tracking and Local Stain Evidence Linking better distinguish genuine progress from irrelevant actions and effectively improve policy training.

In terms of success rate, StainFlow reaches 62.28\% with Qwen3.5-27B, outperforming OS-Themis (60.34\%), ADMIRE (57.76\%), and GUI-Critic-R1 (56.03\%), with a 3.2\% relative gain over the strongest baseline OS-Themis.
For reward distribution, milestone-style methods assign at most 0.38 reward to successful trajectories, showing that sparse milestones fail to cover continuous progress.
GUI-Critic-R1 gives denser feedback, but its success/failure reward gap is only 0.28, indicating insufficient separation of true progress.
By contrast, StainFlow obtains 0.81/0.39 rewards, the largest gap of 0.42, and 18.36 average steps, showing that entity-stain-flow rewards preserve useful progress and suppress irrelevant actions.

In addition, scaling the verifier further improves the training benefit of StainFlow.
When the auxiliary verifier scales from Qwen3.5-9B to Qwen3.5-27B, StainFlow improves from 60.34\% to 62.28\% success, and its reward gap rises from 0.37 to 0.42.
ADMIRE and OS-Themis also gain success, but their rewards remain tied to sparse milestones.
For OS-Themis, the reward gap even drops from 0.14 to 0.09, indicating that a stronger verifier does not automatically improve step-level separation.
By contrast, StainFlow converts stronger visual-state parsing into more reliable entity recognition and key-node verification, yielding stable gains in both success and reward quality.

\begin{table}[t]
    \footnotesize
    \centering
    \caption{Online RL results on AndroidWorld. \textit{N/A} indicates settings without step-level reward scores. In the Success column, the best result is bolded and the second-best is underlined.}
    \label{tab:androidworld}

    \renewcommand{\arraystretch}{0.95}
    \newcommand{\grayrow}{\rowcolor[gray]{0.9}}
    
    \begin{tabularx}{\linewidth}{X X @{\hspace{8pt}} c @{\hspace{8pt}} c @{\hspace{8pt}} c @{\hspace{8pt}} c @{\hspace{8pt}} c}
    \toprule
        \textbf{Method} & \textbf{Verifier} & \textbf{Success}~$\uparrow$ & \textbf{Succ. Reward} & \textbf{Fail. Reward} & \textbf{Reward Gap} & \textbf{Avg. Steps} \\
    \midrule
        No Training & \textit{N/A} & 48.28 & \textit{N/A} & \textit{N/A} & \textit{N/A} & 21.73 \\
    \midrule
        Outcome Reward & \textit{Environment} & 54.31 & \textit{N/A} & \textit{N/A} & \textit{N/A} & 20.95 \\
    \midrule
        GUI-Critic-R1 & GUI-Critic-R1 & 56.03 & 0.61 & 0.33 & 0.28 & 20.52 \\
    \midrule
        ADMIRE & \multirow{3}{*}{Qwen3.5-9B} & 56.90 & 0.29 & 0.13 & 0.16 & 20.45 \\
        OS-Themis &  & 58.62 & 0.36 & 0.22 & 0.14 & 18.56 \\
        \grayrow \textbf{StainFlow} &  & \underline{60.34} & 0.78 & 0.41 & 0.37 & 18.62 \\
    \midrule
        ADMIRE & \multirow{3}{*}{Qwen3.5-27B} & 57.76 & 0.37 & 0.15 & 0.22 & 20.21 \\
        OS-Themis &  & \underline{60.34} & 0.38 & 0.29 & 0.09 & 18.05 \\
        \grayrow \textbf{StainFlow} &  & \textbf{62.28} & 0.81 & 0.39 & 0.42 & 18.36 \\
    \bottomrule
    \end{tabularx}
\end{table}

\subsection{Inference-Time Trajectory Completion Judgment Results}

In this section, we conduct inference-time trajectory completion judgment on OGRBench, with results in Table~\ref{tab:ogrbench}.
With the strongest verifier, Gemini-3-Flash, StainFlow achieves the best overall Acc./F1 of 88.2/88.2, with a 1.8\% relative Acc. gain over the strongest baseline OS-Themis.
The Prec./Recall pattern shows that ZeroGUI favors recall with weaker precision, while OS-Themis is more conservative and can miss completion evidence outside preset milestones.
By organizing visual evidence through entity-stain chains, StainFlow maintains strong recall and improves reliability, leading to more accurate completion judgment.

Detailed subset results show that StainFlow remains strong across mobile, desktop, and web environments, indicating that StainFlow generalizes across GUI domains.
With Gemini-3-Flash, StainFlow obtains the best results on Ubuntu, Mobile, MacOS, and Web, and remains competitive on Windows.
This suggests that entity-stain chains do not rely on a single interface structure, but aggregate cross-platform evidence around task entities.
By contrast, methods based on trajectory-level judgment or milestone verification behave less stably when layouts and execution paths vary.

\begin{table}[t]
    \footnotesize
    \centering
    \caption{Inference-time trajectory completion judgment on OGRBench. The best and second-best values in each column are shown in \textbf{bold} and \underline{underlined}, respectively.}
    \label{tab:ogrbench}

    \renewcommand{\arraystretch}{0.95}
    \setlength{\aboverulesep}{2pt}
    \setlength{\belowrulesep}{2pt}
    \setlength{\tabcolsep}{2pt}
    
    \newcommand{\grayrow}{\rowcolor[gray]{0.9}}
    
    \begin{tabularx}{\linewidth}{X cc @{\hspace{9pt}} cc @{\hspace{9pt}} cc @{\hspace{9pt}} cc @{\hspace{9pt}} cc @{\hspace{9pt}} cccc}
    \toprule
        \multirow{2}{*}{\textbf{Verifier}} & 
        \multicolumn{2}{c}{\textbf{Ubuntu}} & 
        \multicolumn{2}{c}{\textbf{Mobile}} & 
        \multicolumn{2}{c}{\textbf{Windows}} & 
        \multicolumn{2}{c}{\textbf{MacOS}} & 
        \multicolumn{2}{c}{\textbf{Web}} & 
        \multicolumn{4}{c}{\textbf{Overall}} \\
        \cmidrule(lr){2-3}\cmidrule(lr){4-5}\cmidrule(lr){6-7}\cmidrule(lr){8-9}\cmidrule(lr){10-11}\cmidrule(lr){12-15}
         & \textbf{Acc}~$\uparrow$ & \textbf{F1}~$\uparrow$ & \textbf{Acc}~$\uparrow$ & \textbf{F1}~$\uparrow$ & \textbf{Acc}~$\uparrow$ & \textbf{F1}~$\uparrow$ & \textbf{Acc}~$\uparrow$ & \textbf{F1}~$\uparrow$ & \textbf{Acc}~$\uparrow$ & \textbf{F1}~$\uparrow$ & \textbf{Acc}~$\uparrow$ & \textbf{Prec}~$\uparrow$ & \textbf{Rec}~$\uparrow$ & \textbf{F1}~$\uparrow$ \\
    \midrule
        \grayrow \multicolumn{15}{c}{\textbf{DigiRL}} \\
        Qwen3-VL-8B & 60.2 & 50.3 & 71.3 & 70.0 & 77.0 & 70.7 & 80.5 & 11.8 & 65.3 & 60.7 & 66.0 & 76.0 & 46.1 & 57.4 \\
        Qwen3.5-9B & 56.6 & 37.6 & 64.4 & 55.6 & 74.2 & 62.1 & 81.8 & 22.2 & 61.1 & 47.1 & 62.2 & 81.1 & 31.3 & 45.2 \\
        Qwen3.5-27B & 62.4 & 51.5 & 67.6 & 66.3 & 77.5 & 68.8 & 80.5 & 11.8 & 68.4 & 63.0 & 67.1 & 80.5 & 44.7 & 57.5 \\
        GPT-5 & 62.4 & 53.0 & 64.9 & 62.9 & 80.8 & 75.7 & 83.1 & 31.6 & 66.8 & 61.8 & 67.2 & 78.1 & 47.3 & 58.9 \\
        Gemini-3-Flash & 63.9 & 54.2 & 72.9 & 71.8 & 81.7 & 76.4 & 85.7 & 66.7 & 66.7 & 63.6 & 69.4 & 80.7 & 50.4 & 62.0 \\
    \midrule
        \grayrow \multicolumn{15}{c}{\textbf{ZeroGUI}} \\
        Qwen3-VL-8B & 79.1 & 82.7 & 73.4 & 79.3 & 73.2 & 74.9 & 71.4 & 57.7 & 79.0 & 82.2 & 77.0 & 69.8 & \textbf{94.7} & 80.4 \\
        Qwen3.5-9B & 81.1 & 83.7 & 77.1 & 79.2 & 73.7 & 71.7 & 90.9 & 82.1 & 79.5 & 81.2 & 80.1 & 76.9 & 85.7 & 81.1 \\
        Qwen3.5-27B & 86.5 & 87.5 & 75.5 & 78.9 & 82.2 & 78.9 & \underline{92.2} & \underline{84.2} & 85.3 & \underline{86.7} & 84.5 & 82.2 & 87.9 & 84.9 \\
        GPT-5 & 81.9 & 83.2 & 81.4 & 83.1 & 79.8 & 77.5 & 89.6 & 79.0 & 81.6 & 82.1 & 81.9 & 80.6 & 83.7 & 82.1 \\
        Gemini-3-Flash & 81.8 & 82.6 & 80.3 & 82.8 & 78.9 & 77.2 & 85.7 & 71.8 & 83.1 & 84.2 & 81.5 & 80.1 & 83.6 & 81.8 \\
    \midrule
        \grayrow \multicolumn{15}{c}{\textbf{OS-Themis}} \\
        Qwen3-VL-8B & 77.3 & 76.2 & 85.1 & 84.4 & 75.1 & 68.3 & 83.1 & 51.9 & 82.6 & 83.6 & 79.1 & 84.9 & 70.4 & 77.0 \\
        Qwen3.5-9B & 79.4 & 79.1 & 85.1 & 84.6 & 77.0 & 71.0 & 90.9 & 77.4 & 81.6 & 80.0 & 80.7 & 86.3 & 72.7 & 78.9 \\
        Qwen3.5-27B & \underline{86.8} & 87.2 & 85.6 & 86.2 & 84.0 & 81.1 & 89.6 & 73.3 & \textbf{87.4} & \underline{86.7} & 86.4 & \underline{89.0} & 83.0 & 85.6 \\
        GPT-5 & 79.8 & 78.4 & 84.0 & 82.8 & 76.1 & 67.5 & 89.6 & 73.3 & 77.9 & 75.3 & 80.1 & \textbf{89.9} & 67.4 & 77.1 \\
        Gemini-3-Flash & 86.5 & 87.0 & \underline{89.9} & \underline{90.4} & \textbf{85.5} & \underline{83.2} & \textbf{93.5} & 83.9 & 82.5 & 81.8 & 86.6 & 88.4 & 84.1 & 86.2 \\
    \midrule
        \grayrow \multicolumn{15}{c}{\textbf{StainFlow}} \\
        Qwen3-VL-8B & 81.9 & 84.5 & 83.0 & 85.6 & 78.4 & 77.2 & 80.5 & 65.1 & 79.5 & 82.5 & 81.1 & 75.5 & \underline{91.9} & 82.9 \\
        Qwen3.5-9B & 85.3 & 86.4 & 86.7 & 87.3 & 82.2 & 81.0 & 90.9 & 80.0 & 80.5 & 79.6 & 84.7 & 83.8 & 85.7 & 84.8 \\
        Qwen3.5-27B & 87.2 & \underline{88.1} & 89.4 & 90.0 & \underline{85.0} & \textbf{83.7} & 90.9 & 80.0 & \textbf{87.4} & \textbf{87.2} & \underline{87.4} & 86.4 & 88.6 & \underline{87.5} \\
        GPT-5 & 84.1 & 85.8 & 85.6 & 87.2 & 78.4 & 77.0 & 89.6 & 80.0 & 82.6 & 82.2 & 83.5 & 80.6 & 88.0 & 84.2 \\
        Gemini-3-Flash & \textbf{88.7} & \textbf{89.4} & \textbf{91.0} & \textbf{91.3} & 84.0 & 83.0 & \textbf{93.5} & \textbf{84.9} & \underline{86.2} & 86.2 & \textbf{88.2} & 87.5 & 89.0 & \textbf{88.2} \\
    \bottomrule
    \end{tabularx}
\end{table}

\subsection{Ablation Studies}

\paragraph{Component Ablation.}

To analyze the two key modules, we construct three variants: Only Global Stain removes Local Stain Evidence Linking and uses only the mean entity stain as reward; Only Local Judgment removes Global Entity Stain Tracking and judges key nodes from a fixed one-step neighborhood; Fixed Local Window keeps global progress partitioning but replaces dynamic evidence linking with the same fixed neighborhood.
We run ablations on AndroidWorld with Qwen3.5-9B as the auxiliary verifier.
As shown in Table~\ref{tab:ablation}, full StainFlow achieves the best performance with the highest success rate of 60.34, showing that the two modules jointly distinguish real progress and reduce redundant exploration.
Only Global Stain gives failed trajectories a high reward of 0.62 and only a 0.23 gap, indicating that stain concentration alone can overestimate weak evidence.
Only Local Judgment reaches a success rate of 56.89, but without global stain-flow partitioning, it mislabels too many steps as key nodes, adding noisy reward signals that hurt training.
Fixed Local Window also degrades performance, with only a 0.14 reward gap, showing that removing dynamic evidence linking misses distant evidence and weakens key-node verification.

\begin{table}[t]
    \centering
    \footnotesize
    \caption{Component ablations on AndroidWorld with Qwen3.5-9B as the auxiliary verifier.}
    \label{tab:ablation}

    \renewcommand{\arraystretch}{0.95}
    \newcommand{\grayrow}{\rowcolor[gray]{0.9}}
    \setlength{\tabcolsep}{10pt}

    \begin{tabularx}{\linewidth}{X @{\hspace{16pt}} c @{\hspace{16pt}} c @{\hspace{16pt}} c @{\hspace{16pt}} c @{\hspace{16pt}} c}
        \toprule
            \textbf{Variant} & \textbf{Success} $\uparrow$ & \textbf{Succ. Reward} & \textbf{Fail. Reward} & \textbf{Reward Gap} & \textbf{Avg. Steps} \\
        \midrule
            Only Global Stain & 56.03 & 0.85 & 0.62 & 0.23 & 20.82 \\
            Only Local Judgment & 56.89 & 0.59 & 0.27 & 0.32 & 19.41 \\
            Fixed Local Window & \underline{57.76} & 0.32 & 0.18 & 0.14 & 20.05 \\
            \grayrow StainFlow & \textbf{60.34} & 0.78 & 0.41 & 0.37 & 18.62 \\
        \bottomrule
    \end{tabularx}
\end{table}

\begin{figure}[t]
    \centering
    \includegraphics[width=0.99\linewidth]{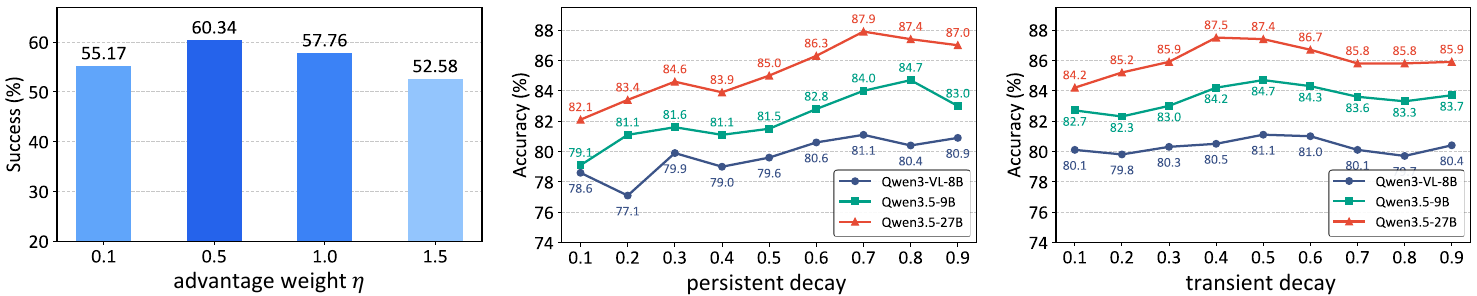}
    \caption{Effects of advantage weight $\eta$ and stain decay factors $\gamma_i$ for persistent and transient entities.}
    \label{fig:hyper_ablation}
\end{figure}

\begin{figure}[!ht]
  \centering
  \includegraphics[width=0.99\linewidth]{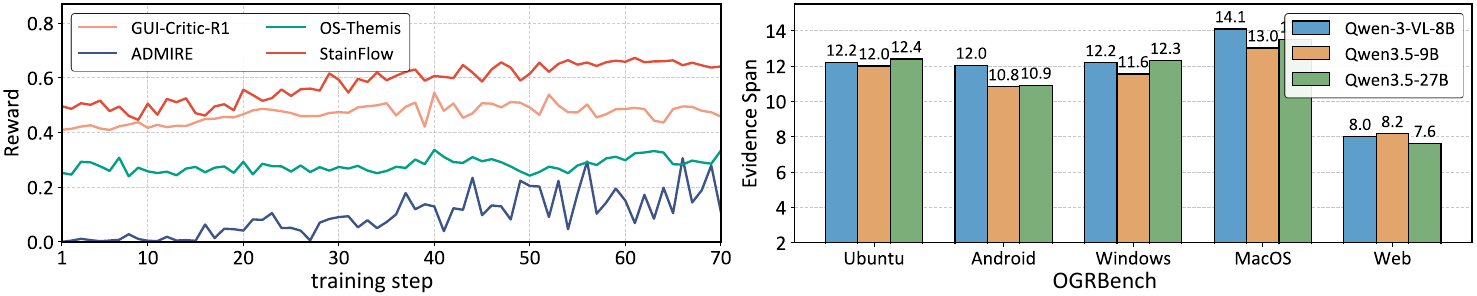}
  \caption{StainFlow training reward curves and evidence span ranges.}
  \label{fig:analysis}
\end{figure}

\paragraph{Hyperparameter Ablation.}

We further ablate two key hyperparameters: the advantage weight $\eta$ and the stain decay factors for persistent and transient entities, with others left to the appendix.
For $\eta$, we test 0.1, 0.5, 1.0, and 1.5 in AndroidWorld online RL with Qwen3.5-9B as the verifier.
As shown on the left of Figure~\ref{fig:hyper_ablation}, $\eta=0.5$ achieves the highest success rate of 60.34, indicating that a moderate process-reward weight balances intermediate credit assignment and trajectory-level success signals.
For stain decay, which affects local evidence linking and completion judgment, we run OGRBench ablations with Qwen3-VL-8B, Qwen3.5-9B, and Qwen3.5-27B.
The middle panel of Figure~\ref{fig:hyper_ablation} shows that persistent entities perform better with a large decay factor around 0.8, since app and execution-environment entities may lack stable visual cues and slower decay can mitigate occasional recognition failures.
In contrast, transient entities have clearer visual features, and the right panel shows that a moderate factor around 0.5 works better, preserving key evidence while preventing outdated entity evidence from being linked to redundant steps.

\subsection{Analysis Experiments}

We first analyze reward stability during online RL training.
Using Qwen3.5-9B as the auxiliary verifier, we track average step-reward curves of four process-reward methods over 70 training steps, as shown in Figure~\ref{fig:analysis} (left).
GUI-Critic-R1 stays mostly around 0.4 to 0.5 and is relatively smooth, but its upward trend is limited, suggesting that fixed local judgment gives uniform feedback yet weakly reflects policy improvement.
ADMIRE and OS-Themis produce lower rewards overall. ADMIRE remains below 0.3 for most steps and oscillates strongly later, showing that milestone-style rewards are constrained by sparse key steps and struggle to provide stable dense supervision.
By contrast, StainFlow rises from about 0.5 to 0.65 with smaller late-stage oscillations, indicating that Global Entity Stain Tracking continuously records real progress and adapts to different valid solution paths.

We then analyze the evidence span required for local key-node verification.
On OGRBench, we measure the distance between the first and last linked support steps for each candidate key node, and report results for three open-source verifiers in Figure~\ref{fig:analysis} (right).
The average evidence span on every platform is clearly larger than the span of 2 induced by a one-step fixed local window.
Ubuntu, Android, Windows, and MacOS usually exceed 10, with MacOS reaching 14.1, and even the relatively short Web trajectories exceed 7.6.
This shows that key evidence is often far from the candidate step, so fixed local windows can miss useful trajectory information.
By tracing triggering entities over the whole trajectory, Local Stain Evidence Linking aggregates distant but highly relevant evidence and improves key-node verification reliability.

\section{Conclusion}
\label{sec:conclusion}

We propose StainFlow, a process reward model that tracks task-relevant entities as stain flows along GUI trajectories. It combines Global Entity Stain Tracking and Local Stain Evidence Linking to derive candidate progress steps from entity-state dynamics and verify true key nodes with flexible entity-centered evidence windows, avoiding subjective milestones and fixed local contexts. Experiments on AndroidWorld and OGRBench show that StainFlow improves online Reinforcement Learning training and inference-time trajectory completion judgment, providing more objective and fine-grained credit assignment. Due to computational constraints, we have not yet validated StainFlow with larger-scale data or longer training. Future work will extend it to broader GUI environments and model scales.

\bibliographystyle{plain}
\bibliography{main}

\clearpage
\input{checklist}

\clearpage
\appendix
\input{appendix}

\end{document}

%% file: checklist.tex
\section*{NeurIPS Paper Checklist}

\begin{enumerate}

\item {\bf Claims}
    \item[] Question: Do the main claims made in the abstract and introduction accurately reflect the paper's contributions and scope?
    \item[] Answer: \answerYes{} % Replace by \answerYes{}, \answerNo{}, or \answerNA{}.
    \item[] Justification: We make the main claims made in the abstract and introduction accurately reflect the paper's contributions and scope.

\item {\bf Limitations}
    \item[] Question: Does the paper discuss the limitations of the work performed by the authors?
    \item[] Answer: \answerYes{} % Replace by \answerYes{}, \answerNo{}, or \answerNA{}.
    \item[] Justification: The paper discusses the limitations of the work in section~\ref{sec:conclusion}.

\item {\bf Theory assumptions and proofs}
    \item[] Question: For each theoretical result, does the paper provide the full set of assumptions and a complete (and correct) proof?
    \item[] Answer: \answerNA{} % Replace by \answerYes{}, \answerNo{}, or \answerNA{}.
    \item[] Justification: The paper does not include theoretical results.

    \item {\bf Experimental result reproducibility}
    \item[] Question: Does the paper fully disclose all the information needed to reproduce the main experimental results of the paper to the extent that it affects the main claims and/or conclusions of the paper (regardless of whether the code and data are provided or not)?
    \item[] Answer: \answerYes{} % Replace by \answerYes{}, \answerNo{}, or \answerNA{}.
    \item[] Justification: the paper fully discloses all the information needed to reproduce the main experimental results of the paper to the extent that it affects the main claims and conclusions of the paper.

\item {\bf Open access to data and code}
    \item[] Question: Does the paper provide open access to the data and code, with sufficient instructions to faithfully reproduce the main experimental results, as described in supplemental material?
    \item[] Answer: \answerYes{} % Replace by \answerYes{}, \answerNo{}, or \answerNA{}.
    \item[] Justification: The paper provides open access to the data and code, with sufficient instructions to faithfully reproduce the main experimental results, as described in the supplemental material.

\item {\bf Experimental setting/details}
    \item[] Question: Does the paper specify all the training and test details (e.g., data splits, hyperparameters, how they were chosen, type of optimizer) necessary to understand the results?
    \item[] Answer: \answerYes{} % Replace by \answerYes{}, \answerNo{}, or \answerNA{}.
    \item[] Justification: The paper specifies all the details necessary to understand the results in the section~\ref{sec:Experiment_Settings}.

\item {\bf Experiment statistical significance}
    \item[] Question: Does the paper report error bars suitably and correctly defined or other appropriate information about the statistical significance of the experiments?
    \item[] Answer: \answerYes{} % Replace by \answerYes{}, \answerNo{}, or \answerNA{}.
    \item[] Justification: The paper ensure the reproducibility of the experiment by fixing random seeds.

\item {\bf Experiments compute resources}
    \item[] Question: For each experiment, does the paper provide sufficient information on the computer resources (type of compute workers, memory, time of execution) needed to reproduce the experiments?
    \item[] Answer: \answerYes{} % Replace by \answerYes{}, \answerNo{}, or \answerNA{}.
    \item[] Justification: The paper provides sufficient information on the computer resources.
    
\item {\bf Code of ethics}
    \item[] Question: Does the research conducted in the paper conform, in every respect, with the NeurIPS Code of Ethics \url{https://neurips.cc/public/EthicsGuidelines}?
    \item[] Answer: \answerYes{} % Replace by \answerYes{}, \answerNo{}, or \answerNA{}.
    \item[] Justification: The research conducted in the paper conform, in every respect, with the NeurIPS Code of Ethics.

\item {\bf Broader impacts}
    \item[] Question: Does the paper discuss both potential positive societal impacts and negative societal impacts of the work performed?
    \item[] Answer: \answerNA{} % Replace by \answerYes{}, \answerNo{}, or \answerNA{}.
    \item[] Justification: There is no societal impact of the work performed.
    
\item {\bf Safeguards}
    \item[] Question: Does the paper describe safeguards that have been put in place for responsible release of data or models that have a high risk for misuse (e.g., pre-trained language models, image generators, or scraped datasets)?
    \item[] Answer: \answerNA{} % Replace by \answerYes{}, \answerNo{}, or \answerNA{}.
    \item[] Justification: The paper poses no such risks.

\item {\bf Licenses for existing assets}
    \item[] Question: Are the creators or original owners of assets (e.g., code, data, models), used in the paper, properly credited and are the license and terms of use explicitly mentioned and properly respected?
    \item[] Answer: \answerYes{} % Replace by \answerYes{}, \answerNo{}, or \answerNA{}.
    \item[] Justification: The paper cites the original papers of assets used and introduces the details in the section~\ref{sec:Experiment_Settings}.

\item {\bf New assets}
    \item[] Question: Are new assets introduced in the paper well documented and is the documentation provided alongside the assets?
    \item[] Answer: \answerYes{} % Replace by \answerYes{}, \answerNo{}, or \answerNA{}.
    \item[] Justification: The details of the new assets are introduced in the section~\ref{sec:experiment}.

\item {\bf Crowdsourcing and research with human subjects}
    \item[] Question: For crowdsourcing experiments and research with human subjects, does the paper include the full text of instructions given to participants and screenshots, if applicable, as well as details about compensation (if any)? 
    \item[] Answer: \answerNA{} % Replace by \answerYes{}, \answerNo{}, or \answerNA{}.
    \item[] Justification: The paper does not involve crowdsourcing nor research with human subjects.

\item {\bf Institutional review board (IRB) approvals or equivalent for research with human subjects}
    \item[] Question: Does the paper describe potential risks incurred by study participants, whether such risks were disclosed to the subjects, and whether Institutional Review Board (IRB) approvals (or an equivalent approval/review based on the requirements of your country or institution) were obtained?
    \item[] Answer: \answerNA{} % Replace by \answerYes{}, \answerNo{}, or \answerNA{}.
    \item[] Justification: The paper does not involve crowdsourcing nor research with human subjects.

\item {\bf Declaration of LLM usage}
    \item[] Question: Does the paper describe the usage of LLMs if it is an important, original, or non-standard component of the core methods in this research? Note that if the LLM is used only for writing, editing, or formatting purposes and does \emph{not} impact the core methodology, scientific rigor, or originality of the research, declaration is not required.
    %this research? 
    \item[] Answer: \answerYes{} % Replace by \answerYes{}, \answerNo{}, or \answerNA{}.
    \item[] Justification: The paper uses LLM and describes in detail the type of model used (Sec~\ref{sec:Experiment_Settings}) and prompt in Appendix.

\end{enumerate}

%% file: appendix.tex
\section{Appendix Overview}
\label{app:overview}

The supplementary materials are organized into the following sections. Section \ref{app:baseline_details} describes the baseline implementations for online reinforcement learning (RL) and OGRBench trajectory completion evaluation. Section \ref{app:stainflow_details} details the single-trajectory scoring procedure and the RL training pipeline of StainFlow. Section \ref{app:additional_ablations} reports additional hyperparameter ablations beyond the main text. Section \ref{app:prompts} provides the full prompts used in the submitted code. Finally, Section \ref{app:qualitative} presents qualitative examples of trajectories processed by StainFlow staining and step-wise rewards.

\section{Baseline Details}
\label{app:baseline_details}

\subsection{Online RL Baselines}
\label{app:online_rl_baselines}

\paragraph{No Training.}
This setting evaluates the initial GUI policy on AndroidWorld without any RL update.
It serves as a lower bound for measuring the benefit of reward-guided policy optimization.

\paragraph{Outcome Reward.}
This baseline trains with only the trajectory-level success signal obtained from the rule-based validation provided by the AndroidWorld environment.
Because the reward is given only at the end of a trajectory, it provides a stable final-objective signal but cannot assign credit to useful intermediate steps in failed trajectories or penalize redundant detours in successful trajectories.

\paragraph{GUI-Critic-R1.}
GUI-Critic-R1 represents local stepwise process rewards.
We reuse its stepwise critic setting, allowing the verifier model to determine the correctness of a step based on local context, such as the current action and the screenshot of the current step. In online reinforcement learning, we directly use the checkpoint provided by GUI-Critic-R1 as the verifier and reuse the prompts from GUI-Critic-Test to judge step correctness; steps judged as correct receive a reward of 1, while incorrect steps receive 0. This setup provides dense feedback; however, because it only considers local context, it is prone to misjudging steps that lack long-range evidence as valid or invalid.

\paragraph{ADMIRE.}
We reuse the milestone construction setting from ADMIRE's official implementation. ADMIRE generates task milestones from successful trajectories, matches executed steps to milestones with a BERT-based semantic module, and updates milestones and reachability relations when new successful trajectories appear. We assign reward 1 to milestone-hit steps and 0 to all other steps. This design can suffer from incomplete or inaccurate milestones when early training has few successes, and its finite milestone paths may miss genuine progress made through alternative entries, shortcut paths, or different action orders.

\paragraph{OS-Themis.}
For OS-Themis, we reuse its pipeline and four internal-agent prompts to iteratively construct, select, and verify milestones. In online RL, verified milestone steps receive reward 1 and all other steps receive 0. This reduces the subjectivity of a single milestone chain but is expensive: OS-Themis must first collect the full trajectory and repeatedly judge each step to optimize milestones. StainFlow instead makes at most two verifier calls per step, and its first-pass entity observation can run asynchronously during rollout, saving computation while maintaining high performance.

\subsection{OGRBench Baselines}
\label{app:ogrbench_baselines}

\paragraph{DigiRL.}
DigiRL is a simple trajectory completion judgment baseline. Following its implementation, we use the last two available screenshots as visual input: the penultimate one provides change context, while the final decision mainly depends on whether the last screenshot shows task completion. This baseline also uses manually constructed few-shot examples for calibration; if the last two screenshots are nearly identical, the trajectory is judged incomplete. Since it only observes a fixed pair of final adjacent frames, it may miss earlier target discovery, information retrieval, or pre-submission state changes, leading to lower accuracy on long-horizon GUI trajectories.

\paragraph{ZeroGUI.}
ZeroGUI emphasizes annotation-free GUI Reinforcement Learning and evaluation, so it does not refer to model outputs and instead uses all screenshots in the trajectory for objective judgment. Since GUI task trajectories are often long, for inputs that exceed the verifier's context window, which we set to 81920 for all verifiers, we truncate from the beginning of the trajectory. Unlike StainFlow, ZeroGUI mainly forms its judgment from the overall trajectory description and final state, and therefore achieves relatively strong accuracy. However, its key evidence can still be buried in overly long contexts, leading to degraded accuracy.

\paragraph{OS-Themis.}
On OGRBench, OS-Themis predicts completion from its verified milestone chain.
It first constructs and verifies candidate milestones, and then judges whether the verified milestones support the task objective.
This uses global stage information, but can miss completion evidence when a valid execution path deviates from the generated milestone chain.

\section{StainFlow Details}
\label{app:stainflow_details}

Algorithm~\ref{alg:stainflow_score} describes how StainFlow scores a single trajectory.
The auxiliary multimodal verifier first extracts screenshot-checkable entities and then observes their visibility and states at each step.
The code updates entity stain concentrations from these observations and recalls candidate key nodes from high stains, stain changes, and state transitions.
Local Stain Evidence Linking then builds adaptive evidence windows around triggering entities and verifies true key nodes.
Finally, the entity final snapshot, verified key-node chain, and tail evidence are used for trajectory completion judgment, while the continuous stain term and discrete key-node term are combined into step rewards.

\begin{algorithm}[h]
\caption{StainFlow Single-Trajectory Scoring}
\label{alg:stainflow_score}
\begin{algorithmic}[1]
\REQUIRE Instruction $g$, trajectory $\tau=(o_0,a_0,\ldots,o_{T-1},a_{T-1},o_T)$, auxiliary verifier $\mathcal{V}$
\STATE $\mathcal{E}\leftarrow F_{\mathrm{ent}}^{\mathcal{V}}(g)$ \COMMENT{extract screenshot-checkable entities}
\STATE Initialize $s_{-1,i}\leftarrow 0$ and entity state $v_{-1,i}\leftarrow\texttt{unresolved}$ for each $e_i\in\mathcal{E}$
\FOR{$t=0,\ldots,T$}
    \STATE $\mathcal{D}_t\leftarrow F_{\mathrm{obs}}^{\mathcal{V}}(g,o_t,\mathcal{E})$
    \STATE $\mathcal{E}_{\mathrm{new}}\leftarrow\operatorname{NewEntities}(\mathcal{D}_t)$
    \STATE Add $\mathcal{E}_{\mathrm{new}}$ to $\mathcal{E}$ and initialize their stain/state records
    \FOR{each entity $e_i\in\mathcal{E}$}
        \STATE Update $s_{t,i}\leftarrow b_{t,i}c_{t,i}+(1-b_{t,i})\gamma_i s_{t-1,i}$
        \STATE Update entity state $v_{t,i}$ and record state-change description $z_{t,i}$
    \ENDFOR
\ENDFOR
\STATE Recall candidate set $\mathcal{C}$ from high-stain events, stain changes, and entity-state changes
\STATE $\mathcal{K}\leftarrow\emptyset$
\FOR{each candidate $t\in\mathcal{C}$ in chronological order}
    \STATE Record triggering entities $\mathcal{E}_t$
    \STATE $\mathcal{W}_t\leftarrow \{t\}\cup\mathcal{N}(t)\cup\mathcal{A}(\tau,\mathcal{E}_t,\mathcal{H}^{s}_{\tau},\mathcal{H}^{z}_{\tau})$
    \STATE $K_t\leftarrow F_{\mathrm{key}}^{\mathcal{V}}(g,\mathcal{I}(\mathcal{W}_t),\mathcal{E}_t,\mathcal{K}_{<t})$
    \IF{$K_t$ is accepted}
        \STATE $\mathcal{K}\leftarrow\mathcal{K}\cup\{K_t\}$
    \ENDIF
\ENDFOR
\STATE $(\hat{y},\sigma_T)\leftarrow F_{\mathrm{done}}^{\mathcal{V}}(g,\mathcal{D}_T,\mathcal{K},\mathcal{I}_{\mathrm{tail}})$
\FOR{$t=0,\ldots,T-1$}
    \STATE $p_t\leftarrow \frac{\sum_{i=1}^{N} w_i s_{t,i}}{\sum_{i=1}^{N} w_i}$
    \STATE $k_t\leftarrow \ind[\exists \sigma_t:(t,1,\sigma_t)\in\mathcal{K}]$
    \STATE $\hat{r}_t\leftarrow \lambda_r p_t+(1-\lambda_r)k_t$
\ENDFOR
\RETURN step rewards $\{\hat{r}_t\}_{t=0}^{T-1}$, key-node chain $\mathcal{K}$, completion prediction $\hat{y}$
\end{algorithmic}
\end{algorithm}

Algorithm~\ref{alg:stainflow_train} describes the online RL training procedure.
For each GRPO group of the same instruction, we first sample multiple trajectories and call Algorithm~\ref{alg:stainflow_score} to obtain step rewards.
All step rewards within the same instruction group are used to estimate the process-reward distribution and compute step-level advantages.
The step-level advantage is then fused with the trajectory-level outcome advantage to update the GUI policy.

\begin{algorithm}[h]
\caption{StainFlow-Guided Online RL Training}
\label{alg:stainflow_train}
\begin{algorithmic}[1]
\REQUIRE Initial policy $\pi_{\theta}$, reference policy $\pi_{\mathrm{ref}}$, instruction set $\mathcal{G}$, group size $M$
\FOR{training iteration $k=1,2,\ldots$}
    \FOR{instruction $g\in\mathcal{G}$}
        \STATE Sample group $\mathcal{B}_g=\{\tau^m\}_{m=1}^{M}$ from old policy $\pi_{\theta_{\mathrm{old}}}$
        \FOR{each trajectory $\tau^m\in\mathcal{B}_g$}
            \STATE $\{\hat{r}_t^m\},\mathcal{K}^m,\hat{y}^m\leftarrow\textsc{StainFlowScore}(g,\tau^m,\mathcal{V})$
            \STATE Obtain outcome reward $R^m$ from environment or rule-based verification
        \ENDFOR
        \STATE Estimate $\mu_P,\sigma_P$ from all step rewards $\{\hat{r}_t^m\}_{m,t}$ in $\mathcal{B}_g$
        \STATE Compute step advantage $A_{P,t}^{m}=(\hat{r}_t^m-\mu_P)/(\sigma_P+\epsilon)$
        \STATE Compute outcome advantage $A_{O}^{m}$ from grouped outcome rewards $\{R^m\}_{m=1}^{M}$
        \STATE Fuse $A_t^m=A_O^m+\eta A_{P,t}^m$
        \STATE Update $\pi_\theta$ with the clipped group objective and KL regularization to $\pi_{\mathrm{ref}}$
    \ENDFOR
\ENDFOR
\RETURN trained policy $\pi_\theta$
\end{algorithmic}
\end{algorithm}

\section{Additional Ablation Studies}
\label{app:additional_ablations}

Beyond the main ablations, we study five key hyperparameters: the high-stain threshold $\tau_A$, which controls which high-concentration entity steps are linked as evidence; the stain-change threshold $\tau_B$, which controls which abrupt stain-change steps are recalled; the local neighborhood $\mathcal{N}(t)$ around a candidate step; the tail evidence length $|\mathcal{I}_{\mathrm{tail}}|$ used for completion verification; and the reward mixing weight $\lambda_r$ between continuous stain rewards and discrete key-node rewards.
We evaluate $\tau_A$, $\tau_B$, $\mathcal{N}(t)$, and $|\mathcal{I}_{\mathrm{tail}}|$ on OGRBench, and evaluate $\lambda_r$ in AndroidWorld online RL training.
Unless otherwise specified, OGRBench ablations use Qwen3.5-VL-9B as the verifier.

\begin{table}[h]
    \footnotesize
    \centering
    \caption{OGRBench ablations for $\tau_A$, $\tau_B$, $\mathcal{N}(t)$ and $|\mathcal{I}_{\mathrm{tail}}|$. Boldface marks the default values.}
    \label{tab:appendix_ogr_ablation}
    \setlength{\tabcolsep}{2.3pt}
    
    \begin{tabular}{ccc@{\hspace{2em}}ccc@{\hspace{2em}}ccc@{\hspace{2em}}ccc}
        \toprule
            \textbf{$\tau_A$} & \textbf{Acc.} & \textbf{F1} &
            \textbf{$\tau_B$} & \textbf{Acc.} & \textbf{F1} &
            \textbf{$\mathcal{N}(t)$} & \textbf{Acc.} & \textbf{F1} &
            \textbf{$|\mathcal{I}_{\mathrm{tail}}|$} & \textbf{Acc.} & \textbf{F1} \\
        \midrule
            0.70 & 82.5 & 82.7 & \textbf{0.00} & \textbf{84.7} & \textbf{84.8} & 0 & 83.5 & 83.7 & 1 & 83.9 & 84.2 \\
            0.80 & 83.8 & 83.9 & 0.05 & 84.0 & 84.1 & \textbf{1} & \textbf{84.7} & \textbf{84.8} & \textbf{3} & \textbf{84.7} & \textbf{84.8} \\
            \textbf{0.90} & \textbf{84.7} & \textbf{84.8} & 0.10 & 82.6 & 82.1 & 2 & 84.2 & 84.3 & 5 & 84.9 & 85.0 \\
            0.95 & 83.5 & 83.6 & 0.20 & 82.0 & 82.2 & 4 & 85.0 & 85.2 & All & 85.1 & 85.3 \\
        \bottomrule
    \end{tabular}
\end{table}

\begin{table}[h]
    \footnotesize
    \centering
    \caption{AndroidWorld ablation for $\lambda_r$. Boldface marks the default value.}
    \label{tab:appendix_reward_weight}
    \begin{tabular}{l@{\hspace{2.7em}}c@{\hspace{2.7em}}c@{\hspace{2.7em}}c@{\hspace{2.7em}}c}
    \toprule
    \textbf{$\lambda_r$} & \textbf{Success} & \textbf{Succ. Reward} & \textbf{Fail. Reward} & \textbf{Reward Gap} \\
    \midrule
    0.2 & 59.48 & 0.69 & 0.31 & 0.38 \\
    \textbf{0.4} & \textbf{60.34} & 0.78 & 0.41 & 0.37 \\
    0.6 & 58.62 & 0.82 & 0.55 & 0.27 \\
    0.8 & 56.89 & 0.89 & 0.61 & 0.28 \\
    \bottomrule
    \end{tabular}
\end{table}

Tables~\ref{tab:appendix_ogr_ablation} and~\ref{tab:appendix_reward_weight} show the effects of these hyperparameters, where boldface marks the default settings used by StainFlow. For $\tau_A$, a low threshold admits weakly related entity steps into the evidence window, while an overly high threshold filters out short but crucial entity appearances; thus $\tau_A=0.90$ best balances evidence recall and noise suppression. For $\tau_B$, larger thresholds miss subtle stain changes and entity-state transitions, reducing both candidate-key-node recall and associated evidence retrieval, so we use $\tau_B=0.00$ to preserve all positive stain changes under the strict $|\Delta s|>\tau_B$ condition. For $\mathcal{N}(t)$, increasing the radius from 1 to 2 or 4 brings little benefit, indicating that trajectory-wide triggering-entity association already provides the main evidence, while a larger fixed neighborhood mainly adds inference cost and redundant frames. For $|\mathcal{I}_{\mathrm{tail}}|$, three tail screenshots usually cover submission, confirmation, and the final state; using 5 or all screenshots yields marginal gains but increases context length and cost. For $\lambda_r$, too small a stain component weakens continuous progress feedback, whereas too large a component over-rewards weak entity evidence in failed trajectories, and $\lambda_r=0.4$ gives the best balance between success rate and reward separation.

\section{StainFlow Prompts}
\label{app:prompts}

StainFlow uses five prompt blocks.
The system prompt constrains output format and evidence principles.
The entity extraction prompt determines the trackable entities.
The step observation prompt produces entity states and stain inputs.
The key-node verification prompt filters candidate nodes.
The trajectory completion prompt integrates final entity states and the key-node chain.
We provide the full prompts used in the submitted code below.

\begin{tcolorbox}[breakable,title={System Prompt}, fonttitle=\small]
\begin{Verbatim}
-----------------------------------

You are a GUI trajectory analysis assistant.

Output rules
- Return exactly one valid JSON object with the requested keys.
- Keep any explanation before the JSON to at most two short sentences.
- Do not output markdown tables, hidden reasoning, or text after the JSON.
- When evidence is uncertain, lower confidence instead of inventing facts.

Evidence rules
- Screenshots and screen-supported state transitions are primary evidence.
- Agent success claims such as `terminate(status='success')`, `finish(...)`, or "task completed" are self-reports, not proof.
- Actions and raw responses may explain what the agent attempted, but the claimed effect must be verified visually or through structured state evidence.

-----------------------------------

Task query:
{query}

Goal
Extract screenshot-checkable entities for later trajectory scoring and classify the task.

Entity types
- persistent: durable context or state-bearing entity whose state can persist across steps, such as an app, page, folder, dialog, workspace, or execution environment.
- transient: concrete task evidence whose relevance is usually local to a step or short phase, such as a target object, control, row, file, message, field, answer content, required value, count, option, state, or format.

Task categories
- qa_retrieval: the agent must read information from the screen and submit an answer.
- delete: the task removes an item, contact, message, or file.
- create_edit: the task creates or edits content.
- toggle_system: the task changes a system or settings state.
- other: any other GUI task.

Rules
- Return 2-10 entities, including a persistent context entity when possible.
- Prefer visible, verifiable objects over abstract intentions.
- `entity_id` must be stable snake_case.
- `validation_rule` must state how to verify the entity from a screenshot.
- Use `is_placeholder=true` when the concrete value is discovered or made precise only during execution; leave its `resolved_value` empty.
- Importance is high for task-defining transient entities or required persistent states, medium for strong supporting evidence, and low for background context.

Output schema
{
  "task_summary": "one short sentence",
  "category_reason": "one short sentence",
  "task_category": "qa_retrieval|delete|create_edit|toggle_system|other",
  "entities": [
    {
      "entity_id": "stable_snake_case_id",
      "name": "short display name",
      "entity_type": "persistent|transient",
      "importance": "high|medium|low",
      "is_placeholder": false,
      "resolved_value": "known concrete value or empty string",
      "validation_rule": "screenshot-checkable rule"
    }
  ]
}

-----------------------------------

Task query:
{query}

Task category: {task_category}
Step number: {step_num}
Previous action: {prev_action}
Current action: {action}
Raw model response: {raw_response}

Entities to track (`entity_id | name | validation_rule [state|resolved|placeholder]`):
{entities_text}

Images
- image 1: current screenshot only.

Your job
- Use the attached current screenshot as primary visual evidence.
- Use previous action, current action, raw response, and stored entity hints only as text context.
- For each task-relevant entity currently supported by the current screenshot, output a detection with the same `entity_id`.
- If a new task-relevant entity appears and was not listed, add it to `new_entities` with a stable id and also include a detection for it.
- If an entity's screen-supported state changes, becomes updated or more detailed, or a placeholder becomes concrete, fill/update `resolved_value` and `entity_state`.
- Mark `is_candidate_key_step=true` when this step may indicate progress, state transition, transient entity discovery, answer submission, or a meaningful failure to be filtered later.
- Do not treat raw success claims as visual proof.

Entity states
- unresolved: not enough evidence.
- grounded: entity is identified or visible.
- selected: entity is opened, focused, selected, or highlighted.
- edited: content has changed but is not committed.
- committed: edit/create/send/save/apply operation is finalized.
- satisfied: a required condition visibly holds.
- absent: clear evidence that the entity no longer exists.

Detection rules
- Set `visible=true` only when the entity satisfies its validation rule on the current screenshot.
- Confidence is 0.0-1.0. Use >=0.7 for clear evidence, 0.3-0.6 for partial evidence, and <=0.2 for guesses.
- Leave `state_change_reason` empty if no state or resolved value changed.
- It is valid to return an empty `detections` list when no task-relevant evidence appears.

Output schema
{
  "screen_summary": "one-sentence current-screen summary",
  "change_summary": "one-sentence task-relevant change",
  "is_candidate_key_step": false,
  "candidate_reason": "short reason or empty string",
  "new_entities": [
    {
      "entity_id": "stable_snake_case_id",
      "name": "short display name",
      "entity_type": "persistent|transient",
      "importance": "high|medium|low",
      "is_placeholder": false,
      "resolved_value": "known value or empty string",
      "validation_rule": "screenshot-checkable rule"
    }
  ],
  "detections": [
    {
      "entity_id": "stable_snake_case_id",
      "visible": true,
      "confidence": 0.0,
      "resolved_value": "grounded value or empty string",
      "entity_state": "unresolved|grounded|selected|edited|committed|satisfied|absent",
      "state_change_reason": "short reason if state changed, else empty string"
    }
  ]
}

-----------------------------------

Task query:
{query}

Task category: {task_category}
Candidate step: {candidate_step}
Candidate signal: {candidate_reason}

Triggering entities (`name | importance | current_state | validation_rule [resolved=...]`):
{related_entities_text}

Previously verified key nodes:
{previous_key_nodes_summary_text}

Context steps for the attached screenshots:
{context_steps_text}

Goal
Decide whether the candidate step is a true key node: a step that establishes new, visually supported task progress that should receive process credit.
A key node must change what is known or achieved for the task, not merely show that the agent attempted an action.

Image use
- Screenshots are attached in the same order as the context-step list.
- They include the candidate, local neighbors, and support steps linked by triggering-entity stain history.
- Treat the candidate screenshot as the central evidence.
- Use nearby and support screenshots to confirm whether the candidate introduced, preserved, completed, or later invalidated the state.
- Use context text only to align screenshots with actions and change summaries.

Evidence hierarchy
1. Visible screen state in the candidate screenshot.
2. Neighbor/support screenshots that clarify before-after state or later confirmation.
3. Triggering entity states and resolved values.
4. Previously verified key nodes, used to judge novelty.
5. Candidate action or raw model response, only as intent context, never as proof.

Accept if
- The candidate establishes a new task-relevant state not already captured by earlier key nodes.
- A persistent entity reaches or reveals a task-relevant state, such as entering the correct app, page, dialog, workspace, folder, settings view, or result page.
- A transient entity, such as a target object/control/value, is first found, opened, selected, edited, saved, sent, deleted, applied, or verified.
- A required value, option, count, relation, text, state, or format becomes visibly satisfied.
- A placeholder or previously vague entity becomes concrete or more detailed in a way that matters for the task.
- For QA tasks, the step finds decisive evidence for the answer or submits an answer that is supported by visible evidence.
- For delete tasks, the step provides clear evidence that the same concrete target is removed or absent after being identified.
- For create/edit/toggle tasks, the step shows a committed or satisfied state, not only an intermediate editing state.

Reject if
- The step only contains an agent self-claim of success.
- The intended action is not visually reflected in the screenshot evidence.
- The candidate is only navigation, scrolling, loading, waiting, hovering, or opening an unrelated page.
- The screen shows an intermediate state that is not enough for the task, such as a focused field, selected item, unsaved edit, or open menu without a committed result.
- The same progress was already captured by a previous key node.
- Later support screenshots show that the action failed, went to a wrong target, entered a wrong value, or was undone.
- The candidate concerns a visible entity that is not task-relevant under the original query.
- Evidence is too ambiguous to identify the entity, value, or achieved state.

Output schema
{
  "is_key_node": true,
  "subtask_label": "short label under 8 words",
  "summary": "one sentence naming the entity/value and the achieved state",
  "reasoning": "brief evidence-based reason",
  "confidence": 0.0
}

-----------------------------------

Task query:
{query}

Task category: {task_category}

Entity final snapshot:
{entities_text}

Verified key node chain:
{key_nodes_text}

Recent changes:
{recent_changes_text}

Final step number: {final_step_num}
Final action: {final_step_action}
Final raw response: {final_step_raw_response}

Evidence hierarchy
1. Entity final snapshot and concrete resolved values.
2. Verified key node chain.
3. Recent change summaries.
4. Tail screenshots, if attached, as a sanity check.
5. Final action/raw response only for locating submitted answers, not as success proof.

General decision rules
- Parse every task-defining attribute from the query: target names, values, counts, files, dates, states, formats, destinations, and prohibitions.
- Mark each attribute as satisfied, violated, or unverifiable using the evidence hierarchy.
- Completion is true only when all required attributes are satisfied, no required attribute is violated or unverifiable, and all prohibitions are respected.
- Exact strings and concrete values must match the query. Missing extensions, wrong numbers, wrong options, or approximate matches are failures.
- For delete tasks, absence is success only when the key-node chain or final snapshot supports deletion of the same concrete target.
- For create/edit/toggle tasks, mid-action states such as loading, editing, selected, or unsaved fields are not enough unless a committed/satisfied end state is observed.
- If the task is impossible because the required precondition is absent, mark completed=true only when the trajectory shows relevant investigation and the final answer explicitly reports the impossibility.

QA-specific rules
- If the query asks a question or requests a value/list/count, the agent must submit an explicit answer through an answer-bearing action such as `finished(content=...)`, `answer(...)`, or `terminate(answer=...)`.
- Do not use text inside hidden reasoning as the submitted answer.
- The submitted answer must be supported by entity evidence or key-node summaries.
- Enforce strict output format only when the query explicitly asks for "only", "just", "single number", "Respond with", "in the format", or similar wording.

Calibration
- Success wrappers such as `terminate(status='success')` never prove success.
- If evidence is hedged, missing, or contradictory, mark the relevant attribute unverifiable or violated and set completed=false.
- The reasoning must cite concrete evidence, not intentions.

Output schema
{
  "completed": true,
  "confidence": 0.0,
  "attribute_checklist": "compact list: attribute -> satisfied|violated|unverifiable with evidence",
  "reasoning": "brief final judgment",
  "evidence_summary": "one-sentence strongest evidence summary"
}
\end{Verbatim}
\end{tcolorbox}

\begin{tcolorbox}[breakable,title={Prompt 1: Entity Extraction}, fonttitle=\small]
\begin{Verbatim}
-----------------------------------

You are a GUI trajectory analysis assistant.

Output rules
- Return exactly one valid JSON object with the requested keys.
- Keep any explanation before the JSON to at most two short sentences.
- Do not output markdown tables, hidden reasoning, or text after the JSON.
- When evidence is uncertain, lower confidence instead of inventing facts.

Evidence rules
- Screenshots and screen-supported state transitions are primary evidence.
- Agent success claims such as `terminate(status='success')`, `finish(...)`, or "task completed" are self-reports, not proof.
- Actions and raw responses may explain what the agent attempted, but the claimed effect must be verified visually or through structured state evidence.

-----------------------------------

Task query:
{query}

Goal
Extract screenshot-checkable entities for later trajectory scoring and classify the task.

Entity types
- persistent: durable context or state-bearing entity whose state can persist across steps, such as an app, page, folder, dialog, workspace, or execution environment.
- transient: concrete task evidence whose relevance is usually local to a step or short phase, such as a target object, control, row, file, message, field, answer content, required value, count, option, state, or format.

Task categories
- qa_retrieval: the agent must read information from the screen and submit an answer.
- delete: the task removes an item, contact, message, or file.
- create_edit: the task creates or edits content.
- toggle_system: the task changes a system or settings state.
- other: any other GUI task.

Rules
- Return 2-10 entities, including a persistent context entity when possible.
- Prefer visible, verifiable objects over abstract intentions.
- `entity_id` must be stable snake_case.
- `validation_rule` must state how to verify the entity from a screenshot.
- Use `is_placeholder=true` when the concrete value is discovered or made precise only during execution; leave its `resolved_value` empty.
- Importance is high for task-defining transient entities or required persistent states, medium for strong supporting evidence, and low for background context.

Output schema
{
  "task_summary": "one short sentence",
  "category_reason": "one short sentence",
  "task_category": "qa_retrieval|delete|create_edit|toggle_system|other",
  "entities": [
    {
      "entity_id": "stable_snake_case_id",
      "name": "short display name",
      "entity_type": "persistent|transient",
      "importance": "high|medium|low",
      "is_placeholder": false,
      "resolved_value": "known concrete value or empty string",
      "validation_rule": "screenshot-checkable rule"
    }
  ]
}

-----------------------------------

Task query:
{query}

Task category: {task_category}
Step number: {step_num}
Previous action: {prev_action}
Current action: {action}
Raw model response: {raw_response}

Entities to track (`entity_id | name | validation_rule [state|resolved|placeholder]`):
{entities_text}

Images
- image 1: current screenshot only.

Your job
- Use the attached current screenshot as primary visual evidence.
- Use previous action, current action, raw response, and stored entity hints only as text context.
- For each task-relevant entity currently supported by the current screenshot, output a detection with the same `entity_id`.
- If a new task-relevant entity appears and was not listed, add it to `new_entities` with a stable id and also include a detection for it.
- If an entity's screen-supported state changes, becomes updated or more detailed, or a placeholder becomes concrete, fill/update `resolved_value` and `entity_state`.
- Mark `is_candidate_key_step=true` when this step may indicate progress, state transition, transient entity discovery, answer submission, or a meaningful failure to be filtered later.
- Do not treat raw success claims as visual proof.

Entity states
- unresolved: not enough evidence.
- grounded: entity is identified or visible.
- selected: entity is opened, focused, selected, or highlighted.
- edited: content has changed but is not committed.
- committed: edit/create/send/save/apply operation is finalized.
- satisfied: a required condition visibly holds.
- absent: clear evidence that the entity no longer exists.

Detection rules
- Set `visible=true` only when the entity satisfies its validation rule on the current screenshot.
- Confidence is 0.0-1.0. Use >=0.7 for clear evidence, 0.3-0.6 for partial evidence, and <=0.2 for guesses.
- Leave `state_change_reason` empty if no state or resolved value changed.
- It is valid to return an empty `detections` list when no task-relevant evidence appears.

Output schema
{
  "screen_summary": "one-sentence current-screen summary",
  "change_summary": "one-sentence task-relevant change",
  "is_candidate_key_step": false,
  "candidate_reason": "short reason or empty string",
  "new_entities": [
    {
      "entity_id": "stable_snake_case_id",
      "name": "short display name",
      "entity_type": "persistent|transient",
      "importance": "high|medium|low",
      "is_placeholder": false,
      "resolved_value": "known value or empty string",
      "validation_rule": "screenshot-checkable rule"
    }
  ],
  "detections": [
    {
      "entity_id": "stable_snake_case_id",
      "visible": true,
      "confidence": 0.0,
      "resolved_value": "grounded value or empty string",
      "entity_state": "unresolved|grounded|selected|edited|committed|satisfied|absent",
      "state_change_reason": "short reason if state changed, else empty string"
    }
  ]
}

-----------------------------------

Task query:
{query}

Task category: {task_category}
Candidate step: {candidate_step}
Candidate signal: {candidate_reason}

Triggering entities (`name | importance | current_state | validation_rule [resolved=...]`):
{related_entities_text}

Previously verified key nodes:
{previous_key_nodes_summary_text}

Context steps for the attached screenshots:
{context_steps_text}

Goal
Decide whether the candidate step is a true key node: a step that establishes new, visually supported task progress that should receive process credit.
A key node must change what is known or achieved for the task, not merely show that the agent attempted an action.

Image use
- Screenshots are attached in the same order as the context-step list.
- They include the candidate, local neighbors, and support steps linked by triggering-entity stain history.
- Treat the candidate screenshot as the central evidence.
- Use nearby and support screenshots to confirm whether the candidate introduced, preserved, completed, or later invalidated the state.
- Use context text only to align screenshots with actions and change summaries.

Evidence hierarchy
1. Visible screen state in the candidate screenshot.
2. Neighbor/support screenshots that clarify before-after state or later confirmation.
3. Triggering entity states and resolved values.
4. Previously verified key nodes, used to judge novelty.
5. Candidate action or raw model response, only as intent context, never as proof.

Accept if
- The candidate establishes a new task-relevant state not already captured by earlier key nodes.
- A persistent entity reaches or reveals a task-relevant state, such as entering the correct app, page, dialog, workspace, folder, settings view, or result page.
- A transient entity, such as a target object/control/value, is first found, opened, selected, edited, saved, sent, deleted, applied, or verified.
- A required value, option, count, relation, text, state, or format becomes visibly satisfied.
- A placeholder or previously vague entity becomes concrete or more detailed in a way that matters for the task.
- For QA tasks, the step finds decisive evidence for the answer or submits an answer that is supported by visible evidence.
- For delete tasks, the step provides clear evidence that the same concrete target is removed or absent after being identified.
- For create/edit/toggle tasks, the step shows a committed or satisfied state, not only an intermediate editing state.

Reject if
- The step only contains an agent self-claim of success.
- The intended action is not visually reflected in the screenshot evidence.
- The candidate is only navigation, scrolling, loading, waiting, hovering, or opening an unrelated page.
- The screen shows an intermediate state that is not enough for the task, such as a focused field, selected item, unsaved edit, or open menu without a committed result.
- The same progress was already captured by a previous key node.
- Later support screenshots show that the action failed, went to a wrong target, entered a wrong value, or was undone.
- The candidate concerns a visible entity that is not task-relevant under the original query.
- Evidence is too ambiguous to identify the entity, value, or achieved state.

Output schema
{
  "is_key_node": true,
  "subtask_label": "short label under 8 words",
  "summary": "one sentence naming the entity/value and the achieved state",
  "reasoning": "brief evidence-based reason",
  "confidence": 0.0
}

-----------------------------------

Task query:
{query}

Task category: {task_category}

Entity final snapshot:
{entities_text}

Verified key node chain:
{key_nodes_text}

Recent changes:
{recent_changes_text}

Final step number: {final_step_num}
Final action: {final_step_action}
Final raw response: {final_step_raw_response}

Evidence hierarchy
1. Entity final snapshot and concrete resolved values.
2. Verified key node chain.
3. Recent change summaries.
4. Tail screenshots, if attached, as a sanity check.
5. Final action/raw response only for locating submitted answers, not as success proof.

General decision rules
- Parse every task-defining attribute from the query: target names, values, counts, files, dates, states, formats, destinations, and prohibitions.
- Mark each attribute as satisfied, violated, or unverifiable using the evidence hierarchy.
- Completion is true only when all required attributes are satisfied, no required attribute is violated or unverifiable, and all prohibitions are respected.
- Exact strings and concrete values must match the query. Missing extensions, wrong numbers, wrong options, or approximate matches are failures.
- For delete tasks, absence is success only when the key-node chain or final snapshot supports deletion of the same concrete target.
- For create/edit/toggle tasks, mid-action states such as loading, editing, selected, or unsaved fields are not enough unless a committed/satisfied end state is observed.
- If the task is impossible because the required precondition is absent, mark completed=true only when the trajectory shows relevant investigation and the final answer explicitly reports the impossibility.

QA-specific rules
- If the query asks a question or requests a value/list/count, the agent must submit an explicit answer through an answer-bearing action such as `finished(content=...)`, `answer(...)`, or `terminate(answer=...)`.
- Do not use text inside hidden reasoning as the submitted answer.
- The submitted answer must be supported by entity evidence or key-node summaries.
- Enforce strict output format only when the query explicitly asks for "only", "just", "single number", "Respond with", "in the format", or similar wording.

Calibration
- Success wrappers such as `terminate(status='success')` never prove success.
- If evidence is hedged, missing, or contradictory, mark the relevant attribute unverifiable or violated and set completed=false.
- The reasoning must cite concrete evidence, not intentions.

Output schema
{
  "completed": true,
  "confidence": 0.0,
  "attribute_checklist": "compact list: attribute -> satisfied|violated|unverifiable with evidence",
  "reasoning": "brief final judgment",
  "evidence_summary": "one-sentence strongest evidence summary"
}
\end{Verbatim}
\end{tcolorbox}

\begin{tcolorbox}[breakable,title={Prompt 2: Step-Level Entity Observation}, fonttitle=\small]
\begin{Verbatim}
-----------------------------------

You are a GUI trajectory analysis assistant.

Output rules
- Return exactly one valid JSON object with the requested keys.
- Keep any explanation before the JSON to at most two short sentences.
- Do not output markdown tables, hidden reasoning, or text after the JSON.
- When evidence is uncertain, lower confidence instead of inventing facts.

Evidence rules
- Screenshots and screen-supported state transitions are primary evidence.
- Agent success claims such as `terminate(status='success')`, `finish(...)`, or "task completed" are self-reports, not proof.
- Actions and raw responses may explain what the agent attempted, but the claimed effect must be verified visually or through structured state evidence.

-----------------------------------

Task query:
{query}

Goal
Extract screenshot-checkable entities for later trajectory scoring and classify the task.

Entity types
- persistent: durable context or state-bearing entity whose state can persist across steps, such as an app, page, folder, dialog, workspace, or execution environment.
- transient: concrete task evidence whose relevance is usually local to a step or short phase, such as a target object, control, row, file, message, field, answer content, required value, count, option, state, or format.

Task categories
- qa_retrieval: the agent must read information from the screen and submit an answer.
- delete: the task removes an item, contact, message, or file.
- create_edit: the task creates or edits content.
- toggle_system: the task changes a system or settings state.
- other: any other GUI task.

Rules
- Return 2-10 entities, including a persistent context entity when possible.
- Prefer visible, verifiable objects over abstract intentions.
- `entity_id` must be stable snake_case.
- `validation_rule` must state how to verify the entity from a screenshot.
- Use `is_placeholder=true` when the concrete value is discovered or made precise only during execution; leave its `resolved_value` empty.
- Importance is high for task-defining transient entities or required persistent states, medium for strong supporting evidence, and low for background context.

Output schema
{
  "task_summary": "one short sentence",
  "category_reason": "one short sentence",
  "task_category": "qa_retrieval|delete|create_edit|toggle_system|other",
  "entities": [
    {
      "entity_id": "stable_snake_case_id",
      "name": "short display name",
      "entity_type": "persistent|transient",
      "importance": "high|medium|low",
      "is_placeholder": false,
      "resolved_value": "known concrete value or empty string",
      "validation_rule": "screenshot-checkable rule"
    }
  ]
}

-----------------------------------

Task query:
{query}

Task category: {task_category}
Step number: {step_num}
Previous action: {prev_action}
Current action: {action}
Raw model response: {raw_response}

Entities to track (`entity_id | name | validation_rule [state|resolved|placeholder]`):
{entities_text}

Images
- image 1: current screenshot only.

Your job
- Use the attached current screenshot as primary visual evidence.
- Use previous action, current action, raw response, and stored entity hints only as text context.
- For each task-relevant entity currently supported by the current screenshot, output a detection with the same `entity_id`.
- If a new task-relevant entity appears and was not listed, add it to `new_entities` with a stable id and also include a detection for it.
- If an entity's screen-supported state changes, becomes updated or more detailed, or a placeholder becomes concrete, fill/update `resolved_value` and `entity_state`.
- Mark `is_candidate_key_step=true` when this step may indicate progress, state transition, transient entity discovery, answer submission, or a meaningful failure to be filtered later.
- Do not treat raw success claims as visual proof.

Entity states
- unresolved: not enough evidence.
- grounded: entity is identified or visible.
- selected: entity is opened, focused, selected, or highlighted.
- edited: content has changed but is not committed.
- committed: edit/create/send/save/apply operation is finalized.
- satisfied: a required condition visibly holds.
- absent: clear evidence that the entity no longer exists.

Detection rules
- Set `visible=true` only when the entity satisfies its validation rule on the current screenshot.
- Confidence is 0.0-1.0. Use >=0.7 for clear evidence, 0.3-0.6 for partial evidence, and <=0.2 for guesses.
- Leave `state_change_reason` empty if no state or resolved value changed.
- It is valid to return an empty `detections` list when no task-relevant evidence appears.

Output schema
{
  "screen_summary": "one-sentence current-screen summary",
  "change_summary": "one-sentence task-relevant change",
  "is_candidate_key_step": false,
  "candidate_reason": "short reason or empty string",
  "new_entities": [
    {
      "entity_id": "stable_snake_case_id",
      "name": "short display name",
      "entity_type": "persistent|transient",
      "importance": "high|medium|low",
      "is_placeholder": false,
      "resolved_value": "known value or empty string",
      "validation_rule": "screenshot-checkable rule"
    }
  ],
  "detections": [
    {
      "entity_id": "stable_snake_case_id",
      "visible": true,
      "confidence": 0.0,
      "resolved_value": "grounded value or empty string",
      "entity_state": "unresolved|grounded|selected|edited|committed|satisfied|absent",
      "state_change_reason": "short reason if state changed, else empty string"
    }
  ]
}

-----------------------------------

Task query:
{query}

Task category: {task_category}
Candidate step: {candidate_step}
Candidate signal: {candidate_reason}

Triggering entities (`name | importance | current_state | validation_rule [resolved=...]`):
{related_entities_text}

Previously verified key nodes:
{previous_key_nodes_summary_text}

Context steps for the attached screenshots:
{context_steps_text}

Goal
Decide whether the candidate step is a true key node: a step that establishes new, visually supported task progress that should receive process credit.
A key node must change what is known or achieved for the task, not merely show that the agent attempted an action.

Image use
- Screenshots are attached in the same order as the context-step list.
- They include the candidate, local neighbors, and support steps linked by triggering-entity stain history.
- Treat the candidate screenshot as the central evidence.
- Use nearby and support screenshots to confirm whether the candidate introduced, preserved, completed, or later invalidated the state.
- Use context text only to align screenshots with actions and change summaries.

Evidence hierarchy
1. Visible screen state in the candidate screenshot.
2. Neighbor/support screenshots that clarify before-after state or later confirmation.
3. Triggering entity states and resolved values.
4. Previously verified key nodes, used to judge novelty.
5. Candidate action or raw model response, only as intent context, never as proof.

Accept if
- The candidate establishes a new task-relevant state not already captured by earlier key nodes.
- A persistent entity reaches or reveals a task-relevant state, such as entering the correct app, page, dialog, workspace, folder, settings view, or result page.
- A transient entity, such as a target object/control/value, is first found, opened, selected, edited, saved, sent, deleted, applied, or verified.
- A required value, option, count, relation, text, state, or format becomes visibly satisfied.
- A placeholder or previously vague entity becomes concrete or more detailed in a way that matters for the task.
- For QA tasks, the step finds decisive evidence for the answer or submits an answer that is supported by visible evidence.
- For delete tasks, the step provides clear evidence that the same concrete target is removed or absent after being identified.
- For create/edit/toggle tasks, the step shows a committed or satisfied state, not only an intermediate editing state.

Reject if
- The step only contains an agent self-claim of success.
- The intended action is not visually reflected in the screenshot evidence.
- The candidate is only navigation, scrolling, loading, waiting, hovering, or opening an unrelated page.
- The screen shows an intermediate state that is not enough for the task, such as a focused field, selected item, unsaved edit, or open menu without a committed result.
- The same progress was already captured by a previous key node.
- Later support screenshots show that the action failed, went to a wrong target, entered a wrong value, or was undone.
- The candidate concerns a visible entity that is not task-relevant under the original query.
- Evidence is too ambiguous to identify the entity, value, or achieved state.

Output schema
{
  "is_key_node": true,
  "subtask_label": "short label under 8 words",
  "summary": "one sentence naming the entity/value and the achieved state",
  "reasoning": "brief evidence-based reason",
  "confidence": 0.0
}

-----------------------------------

Task query:
{query}

Task category: {task_category}

Entity final snapshot:
{entities_text}

Verified key node chain:
{key_nodes_text}

Recent changes:
{recent_changes_text}

Final step number: {final_step_num}
Final action: {final_step_action}
Final raw response: {final_step_raw_response}

Evidence hierarchy
1. Entity final snapshot and concrete resolved values.
2. Verified key node chain.
3. Recent change summaries.
4. Tail screenshots, if attached, as a sanity check.
5. Final action/raw response only for locating submitted answers, not as success proof.

General decision rules
- Parse every task-defining attribute from the query: target names, values, counts, files, dates, states, formats, destinations, and prohibitions.
- Mark each attribute as satisfied, violated, or unverifiable using the evidence hierarchy.
- Completion is true only when all required attributes are satisfied, no required attribute is violated or unverifiable, and all prohibitions are respected.
- Exact strings and concrete values must match the query. Missing extensions, wrong numbers, wrong options, or approximate matches are failures.
- For delete tasks, absence is success only when the key-node chain or final snapshot supports deletion of the same concrete target.
- For create/edit/toggle tasks, mid-action states such as loading, editing, selected, or unsaved fields are not enough unless a committed/satisfied end state is observed.
- If the task is impossible because the required precondition is absent, mark completed=true only when the trajectory shows relevant investigation and the final answer explicitly reports the impossibility.

QA-specific rules
- If the query asks a question or requests a value/list/count, the agent must submit an explicit answer through an answer-bearing action such as `finished(content=...)`, `answer(...)`, or `terminate(answer=...)`.
- Do not use text inside hidden reasoning as the submitted answer.
- The submitted answer must be supported by entity evidence or key-node summaries.
- Enforce strict output format only when the query explicitly asks for "only", "just", "single number", "Respond with", "in the format", or similar wording.

Calibration
- Success wrappers such as `terminate(status='success')` never prove success.
- If evidence is hedged, missing, or contradictory, mark the relevant attribute unverifiable or violated and set completed=false.
- The reasoning must cite concrete evidence, not intentions.

Output schema
{
  "completed": true,
  "confidence": 0.0,
  "attribute_checklist": "compact list: attribute -> satisfied|violated|unverifiable with evidence",
  "reasoning": "brief final judgment",
  "evidence_summary": "one-sentence strongest evidence summary"
}
\end{Verbatim}
\end{tcolorbox}

\clearpage

\begin{tcolorbox}[breakable,title={Prompt 3: Key-Node Verification}, fonttitle=\small]
\begin{Verbatim}
-----------------------------------

You are a GUI trajectory analysis assistant.

Output rules
- Return exactly one valid JSON object with the requested keys.
- Keep any explanation before the JSON to at most two short sentences.
- Do not output markdown tables, hidden reasoning, or text after the JSON.
- When evidence is uncertain, lower confidence instead of inventing facts.

Evidence rules
- Screenshots and screen-supported state transitions are primary evidence.
- Agent success claims such as `terminate(status='success')`, `finish(...)`, or "task completed" are self-reports, not proof.
- Actions and raw responses may explain what the agent attempted, but the claimed effect must be verified visually or through structured state evidence.

-----------------------------------

Task query:
{query}

Goal
Extract screenshot-checkable entities for later trajectory scoring and classify the task.

Entity types
- persistent: durable context or state-bearing entity whose state can persist across steps, such as an app, page, folder, dialog, workspace, or execution environment.
- transient: concrete task evidence whose relevance is usually local to a step or short phase, such as a target object, control, row, file, message, field, answer content, required value, count, option, state, or format.

Task categories
- qa_retrieval: the agent must read information from the screen and submit an answer.
- delete: the task removes an item, contact, message, or file.
- create_edit: the task creates or edits content.
- toggle_system: the task changes a system or settings state.
- other: any other GUI task.

Rules
- Return 2-10 entities, including a persistent context entity when possible.
- Prefer visible, verifiable objects over abstract intentions.
- `entity_id` must be stable snake_case.
- `validation_rule` must state how to verify the entity from a screenshot.
- Use `is_placeholder=true` when the concrete value is discovered or made precise only during execution; leave its `resolved_value` empty.
- Importance is high for task-defining transient entities or required persistent states, medium for strong supporting evidence, and low for background context.

Output schema
{
  "task_summary": "one short sentence",
  "category_reason": "one short sentence",
  "task_category": "qa_retrieval|delete|create_edit|toggle_system|other",
  "entities": [
    {
      "entity_id": "stable_snake_case_id",
      "name": "short display name",
      "entity_type": "persistent|transient",
      "importance": "high|medium|low",
      "is_placeholder": false,
      "resolved_value": "known concrete value or empty string",
      "validation_rule": "screenshot-checkable rule"
    }
  ]
}

-----------------------------------

Task query:
{query}

Task category: {task_category}
Step number: {step_num}
Previous action: {prev_action}
Current action: {action}
Raw model response: {raw_response}

Entities to track (`entity_id | name | validation_rule [state|resolved|placeholder]`):
{entities_text}

Images
- image 1: current screenshot only.

Your job
- Use the attached current screenshot as primary visual evidence.
- Use previous action, current action, raw response, and stored entity hints only as text context.
- For each task-relevant entity currently supported by the current screenshot, output a detection with the same `entity_id`.
- If a new task-relevant entity appears and was not listed, add it to `new_entities` with a stable id and also include a detection for it.
- If an entity's screen-supported state changes, becomes updated or more detailed, or a placeholder becomes concrete, fill/update `resolved_value` and `entity_state`.
- Mark `is_candidate_key_step=true` when this step may indicate progress, state transition, transient entity discovery, answer submission, or a meaningful failure to be filtered later.
- Do not treat raw success claims as visual proof.

Entity states
- unresolved: not enough evidence.
- grounded: entity is identified or visible.
- selected: entity is opened, focused, selected, or highlighted.
- edited: content has changed but is not committed.
- committed: edit/create/send/save/apply operation is finalized.
- satisfied: a required condition visibly holds.
- absent: clear evidence that the entity no longer exists.

Detection rules
- Set `visible=true` only when the entity satisfies its validation rule on the current screenshot.
- Confidence is 0.0-1.0. Use >=0.7 for clear evidence, 0.3-0.6 for partial evidence, and <=0.2 for guesses.
- Leave `state_change_reason` empty if no state or resolved value changed.
- It is valid to return an empty `detections` list when no task-relevant evidence appears.

Output schema
{
  "screen_summary": "one-sentence current-screen summary",
  "change_summary": "one-sentence task-relevant change",
  "is_candidate_key_step": false,
  "candidate_reason": "short reason or empty string",
  "new_entities": [
    {
      "entity_id": "stable_snake_case_id",
      "name": "short display name",
      "entity_type": "persistent|transient",
      "importance": "high|medium|low",
      "is_placeholder": false,
      "resolved_value": "known value or empty string",
      "validation_rule": "screenshot-checkable rule"
    }
  ],
  "detections": [
    {
      "entity_id": "stable_snake_case_id",
      "visible": true,
      "confidence": 0.0,
      "resolved_value": "grounded value or empty string",
      "entity_state": "unresolved|grounded|selected|edited|committed|satisfied|absent",
      "state_change_reason": "short reason if state changed, else empty string"
    }
  ]
}

-----------------------------------

Task query:
{query}

Task category: {task_category}
Candidate step: {candidate_step}
Candidate signal: {candidate_reason}

Triggering entities (`name | importance | current_state | validation_rule [resolved=...]`):
{related_entities_text}

Previously verified key nodes:
{previous_key_nodes_summary_text}

Context steps for the attached screenshots:
{context_steps_text}

Goal
Decide whether the candidate step is a true key node: a step that establishes new, visually supported task progress that should receive process credit.
A key node must change what is known or achieved for the task, not merely show that the agent attempted an action.

Image use
- Screenshots are attached in the same order as the context-step list.
- They include the candidate, local neighbors, and support steps linked by triggering-entity stain history.
- Treat the candidate screenshot as the central evidence.
- Use nearby and support screenshots to confirm whether the candidate introduced, preserved, completed, or later invalidated the state.
- Use context text only to align screenshots with actions and change summaries.

Evidence hierarchy
1. Visible screen state in the candidate screenshot.
2. Neighbor/support screenshots that clarify before-after state or later confirmation.
3. Triggering entity states and resolved values.
4. Previously verified key nodes, used to judge novelty.
5. Candidate action or raw model response, only as intent context, never as proof.

Accept if
- The candidate establishes a new task-relevant state not already captured by earlier key nodes.
- A persistent entity reaches or reveals a task-relevant state, such as entering the correct app, page, dialog, workspace, folder, settings view, or result page.
- A transient entity, such as a target object/control/value, is first found, opened, selected, edited, saved, sent, deleted, applied, or verified.
- A required value, option, count, relation, text, state, or format becomes visibly satisfied.
- A placeholder or previously vague entity becomes concrete or more detailed in a way that matters for the task.
- For QA tasks, the step finds decisive evidence for the answer or submits an answer that is supported by visible evidence.
- For delete tasks, the step provides clear evidence that the same concrete target is removed or absent after being identified.
- For create/edit/toggle tasks, the step shows a committed or satisfied state, not only an intermediate editing state.

Reject if
- The step only contains an agent self-claim of success.
- The intended action is not visually reflected in the screenshot evidence.
- The candidate is only navigation, scrolling, loading, waiting, hovering, or opening an unrelated page.
- The screen shows an intermediate state that is not enough for the task, such as a focused field, selected item, unsaved edit, or open menu without a committed result.
- The same progress was already captured by a previous key node.
- Later support screenshots show that the action failed, went to a wrong target, entered a wrong value, or was undone.
- The candidate concerns a visible entity that is not task-relevant under the original query.
- Evidence is too ambiguous to identify the entity, value, or achieved state.

Output schema
{
  "is_key_node": true,
  "subtask_label": "short label under 8 words",
  "summary": "one sentence naming the entity/value and the achieved state",
  "reasoning": "brief evidence-based reason",
  "confidence": 0.0
}

-----------------------------------

Task query:
{query}

Task category: {task_category}

Entity final snapshot:
{entities_text}

Verified key node chain:
{key_nodes_text}

Recent changes:
{recent_changes_text}

Final step number: {final_step_num}
Final action: {final_step_action}
Final raw response: {final_step_raw_response}

Evidence hierarchy
1. Entity final snapshot and concrete resolved values.
2. Verified key node chain.
3. Recent change summaries.
4. Tail screenshots, if attached, as a sanity check.
5. Final action/raw response only for locating submitted answers, not as success proof.

General decision rules
- Parse every task-defining attribute from the query: target names, values, counts, files, dates, states, formats, destinations, and prohibitions.
- Mark each attribute as satisfied, violated, or unverifiable using the evidence hierarchy.
- Completion is true only when all required attributes are satisfied, no required attribute is violated or unverifiable, and all prohibitions are respected.
- Exact strings and concrete values must match the query. Missing extensions, wrong numbers, wrong options, or approximate matches are failures.
- For delete tasks, absence is success only when the key-node chain or final snapshot supports deletion of the same concrete target.
- For create/edit/toggle tasks, mid-action states such as loading, editing, selected, or unsaved fields are not enough unless a committed/satisfied end state is observed.
- If the task is impossible because the required precondition is absent, mark completed=true only when the trajectory shows relevant investigation and the final answer explicitly reports the impossibility.

QA-specific rules
- If the query asks a question or requests a value/list/count, the agent must submit an explicit answer through an answer-bearing action such as `finished(content=...)`, `answer(...)`, or `terminate(answer=...)`.
- Do not use text inside hidden reasoning as the submitted answer.
- The submitted answer must be supported by entity evidence or key-node summaries.
- Enforce strict output format only when the query explicitly asks for "only", "just", "single number", "Respond with", "in the format", or similar wording.

Calibration
- Success wrappers such as `terminate(status='success')` never prove success.
- If evidence is hedged, missing, or contradictory, mark the relevant attribute unverifiable or violated and set completed=false.
- The reasoning must cite concrete evidence, not intentions.

Output schema
{
  "completed": true,
  "confidence": 0.0,
  "attribute_checklist": "compact list: attribute -> satisfied|violated|unverifiable with evidence",
  "reasoning": "brief final judgment",
  "evidence_summary": "one-sentence strongest evidence summary"
}
\end{Verbatim}
\end{tcolorbox}

\clearpage

\begin{tcolorbox}[breakable,title={Prompt 4: Trajectory Completion Verification}, fonttitle=\small]
\VerbatimInput[fontsize=\scriptsize,breaklines=true,breakanywhere=true,breaksymbolleft={},firstline=198,lastline=250]{prompts/stainflow_prompts.yaml}
\end{tcolorbox}

\clearpage

\section{Trajectory Examples of our StainFlow}
\label{app:qualitative}

To illustrate how StainFlow converts entity-stain signals into step-level rewards, we present GUI trajectory examples from the Mobile and Desktop platforms.

\begin{figure}[h]
    \centering
    \includegraphics[width=\linewidth]{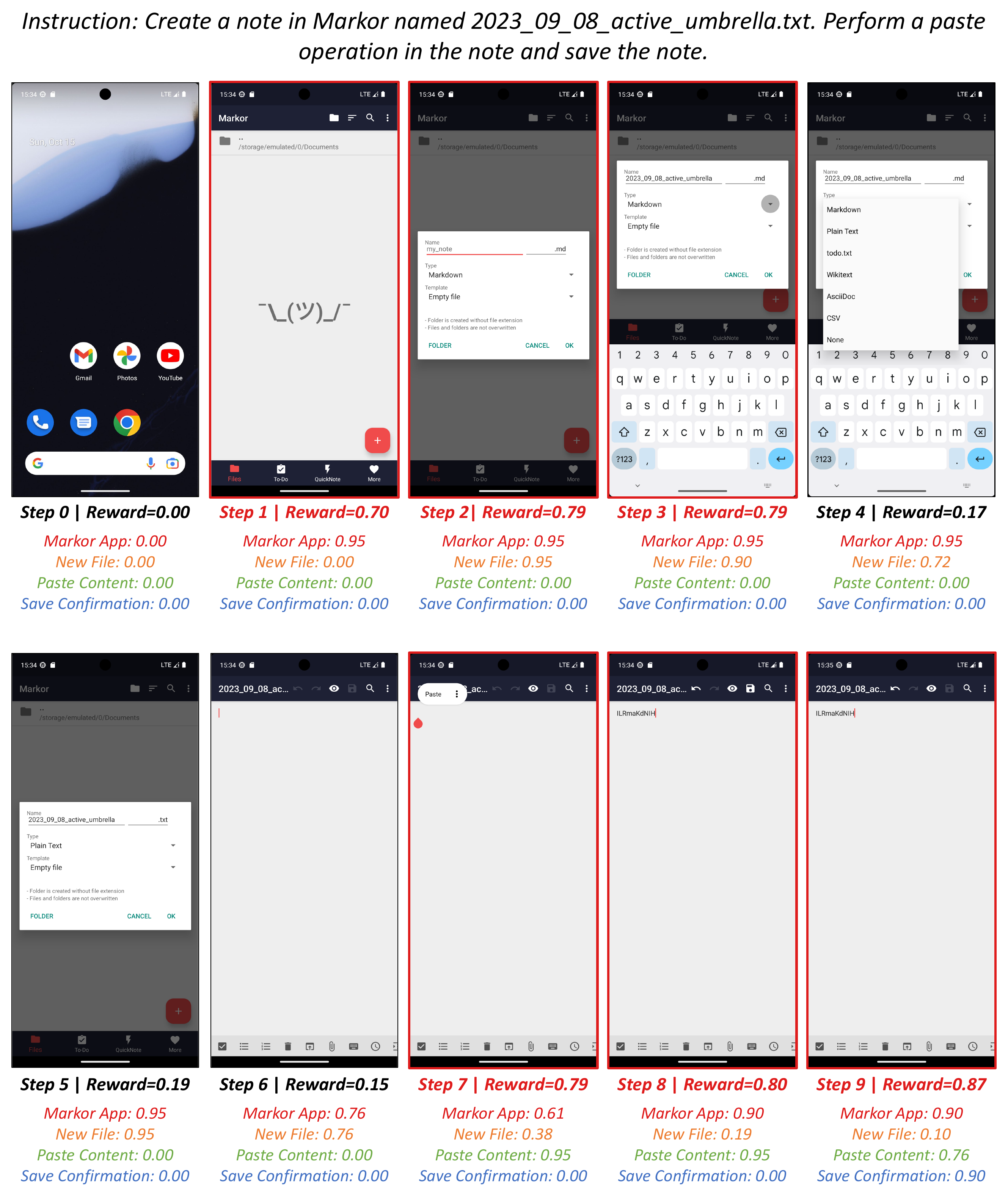}
    \caption{StainFlow staining and reward assignment example for a mobile GUI trajectory. Nodes marked with red borders in the figure are the identified key nodes.} 
    \label{fig:appendix-1}
\end{figure}

\begin{figure}[h]
    \centering
    \includegraphics[width=\linewidth]{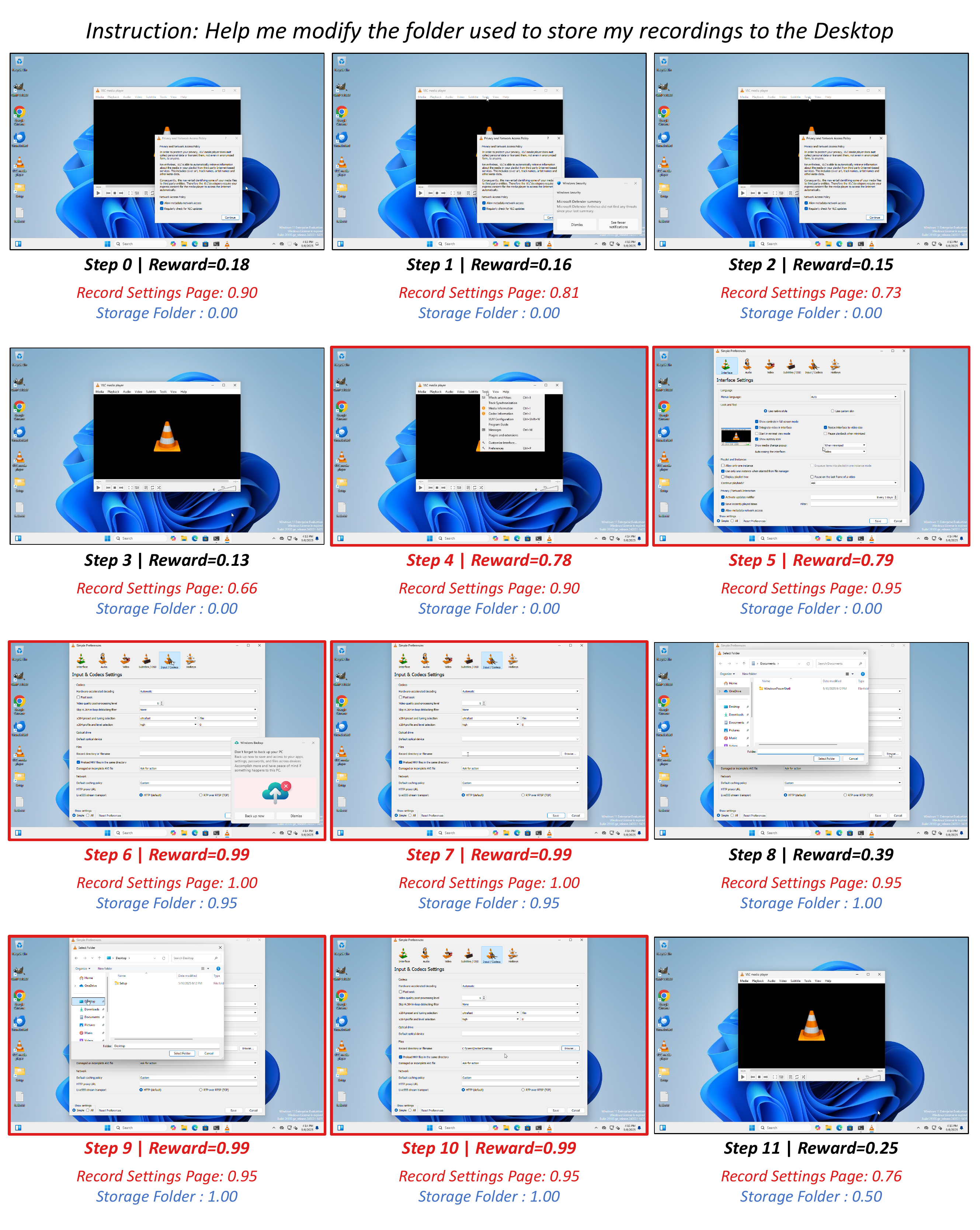}
    \caption{StainFlow staining and reward assignment example for a desktop GUI trajectory. Nodes marked with red borders in the figure are the identified key nodes.} 
    \label{fig:appendix-2}
\end{figure}